\documentclass{article} % For LaTeX2e
\usepackage{iclr2025_conference,times}

\usepackage{amsmath,amsfonts,bm}

\def\eqref#1{equation~\ref{#1}}
\def\1{\bm{1}}

\def\vx{{\bm{x}}}

\def\vz{{\bm{z}}}

\DeclareMathAlphabet{\mathsfit}{\encodingdefault}{\sfdefault}{m}{sl}
\SetMathAlphabet{\mathsfit}{bold}{\encodingdefault}{\sfdefault}{bx}{n}

\def\bepsilon{{\bm{\epsilon}}}

\usepackage{hyperref}
\usepackage{url}

\usepackage{makecell}
\usepackage{multirow}
\usepackage{graphicx}
\usepackage{mathrsfs}
\usepackage{overpic}
\usepackage{pifont}
\newcommand{\cmark}{\ding{51}}%
\newcommand{\xmark}{\ding{55}}%
\newcommand{\rebuttal}[1]{#1}
\newcommand{\modifyCR}[1]{#1}
\usepackage{arydshln}

\usepackage{diagbox}

\def\eg{\emph{e.g.}}

\usepackage{enumitem}
\setlist{itemsep=0pt}
\usepackage{ifthen}

\newif\ifhighresolution
\newif\ifonepdf
\onepdftrue % render the main paper and appendix in one pdf

\ifonepdf
\newcommand{\appendixid}[1]{%
  \ifthenelse{\equal{#1}{shifting_timestep}}{\ref{appendix_transition}}{%
  \ifthenelse{\equal{#1}{experimental_setting}}{\ref{appendix_setting}}{%
  \ifthenelse{\equal{#1}{more_visual_results}}{\ref{appendix_visual}}{%
  \ifthenelse{\equal{#1}{sd3}}{\ref{appendix_sd3}}{%
  \ifthenelse{\equal{#1}{inference_schedule}}{\ref{appendix_ab}}{%
  \ifthenelse{\equal{#1}{each_component}}{\ref{appendix_ab}}{%
  \ifthenelse{\equal{#1}{more_metrics}}{\ref{more_metrics}}{%
  \ifthenelse{\equal{#1}{training_SR}}{\ref{training_SR}}{%
  \textcolor{red}{Warning: no matched id}}}}}}}}}}
\else
\newcommand{\appendixid}[1]{%
  \ifthenelse{\equal{#1}{shifting_timestep}}{A}{%
  \ifthenelse{\equal{#1}{experimental_setting}}{B}{%
  \ifthenelse{\equal{#1}{more_visual_results}}{E}{%
  \ifthenelse{\equal{#1}{sd3}}{F}{%
  \ifthenelse{\equal{#1}{inference_schedule}}{G}{%
  \ifthenelse{\equal{#1}{each_component}}{G}{%
  \ifthenelse{\equal{#1}{more_metrics}}{C}{%
  \ifthenelse{\equal{#1}{training_SR}}{D}{%
  \textcolor{red}{Warning: no matched id}}}}}}}}}}
\fi

\newcommand{\best}[1]{\textbf{\textcolor{red}{#1}}}
\newcommand{\second}[1]{\textcolor{blue}{#1}}

\ifhighresolution
    \graphicspath{{figures}}
\else
    \graphicspath{{figures_smallsize/}}
\fi

\title{FreCaS: Efficient Higher-Resolution Image \\ Generation via Frequency-aware \\ Cascaded Sampling}

\author{Zhengqiang~Zhang$^{1,2}$, Ruihuang~Li$^{1}$, Lei Zhang$^{1,2,^{\dagger}}$\\
	{$^{1}$The Hong Kong Polytechnic University \qquad $^{2}$OPPO Research Institute} \\
	\texttt{zhengqiang.zhang@connect.polyu.hk, cslzhang@comp.polyu.edu.hk}\\
	{$^{\dagger}$Corresponding author}\\
	\ifhighresolution
	\else
	    \\
	    \it{\textcolor{cyan}{For the version with uncompressed images: \href{https://github.com/xtudbxk/FreCaS/FreCaS_high.pdf}{https://github.com/xtudbxk/FreCaS/FreCaS-high.pdf}.}}
	\fi
}

\iclrfinalcopy % Uncomment for camera-ready version, but NOT for submission.
\begin{document}

\maketitle

\begin{abstract}
While image generation with diffusion models has achieved a great success, generating images of higher resolution than the training size remains a challenging task due to the high computational cost.
Current methods typically perform the entire sampling process at full resolution and process all frequency components simultaneously, contradicting with the inherent coarse-to-fine nature of latent diffusion models and wasting computations on processing premature high-frequency details at early diffusion stages.
To address this issue, we introduce an efficient \textbf{Fre}quency-aware \textbf{Ca}scaded \textbf{S}ampling framework, \textbf{FreCaS} in short, for higher-resolution image generation.
FreCaS decomposes the sampling process into cascaded stages with gradually increased resolutions, progressively expanding frequency bands and refining the corresponding details.
We propose an innovative frequency-aware classifier-free guidance (FA-CFG) strategy to assign different guidance strengths for different frequency components, directing the diffusion model to add new details in the expanded frequency domain of each stage.
Additionally, we fuse the cross-attention maps of previous and current stages to avoid synthesizing unfaithful layouts.
Experiments demonstrate that FreCaS significantly outperforms state-of-the-art methods in image quality and generation speed.
In particular, FreCaS is about 2.86$\times$ and 6.07$\times$ faster than ScaleCrafter and DemoFusion in generating a 2048$\times$2048 image using a pre-trained SDXL model and achieves an $\text{FID}_b$ improvement of 11.6 and 3.7, respectively. FreCaS can be easily extended to more complex models such as SD3.
The source code of FreCaS can be found at \url{https://github.com/xtudbxk/FreCaS}.
\end{abstract}

\section{Introducation}
In recent years, diffusion models, such as Imagen~\citep{imagen}, %LDM~\citep{ldm},
SDXL~\citep{sdxl}, PixelArt-$\alpha$~\citep{pixart} and SD3~\cite{sd3}, have achieved a remarkable success in generating high-quality natural images.
However, these models face challenges in generating  very high resolution images due to the increased complexity in high-dimensional space.
Though efficient diffusion models, including  ADM~\citep{adm}, CascadedDM~\citep{cascadeddm} and LDM~\citep{ldm}, have been developed, the computational burden of training diffusion models from scratch for high-resolution image generation remains substantial.
As a result, popular diffusion models, such as SDXL~\citep{sdxl} and SD3~\citep{sd3}, primarily focus on generating $1024 \times 1024$ resolution images.
It is thus increasingly attractive to explore training-free strategies for generating images at higher resolutions, such as $2048 \times 2048$ and $4096 \times 4096$, using pre-trained diffusion models.

MultiDiffusion~\citep{multidiffusion} is among the first works to synthesize higher-resolution images using pre-trained diffusion models. However, it suffers from issues such as object duplication, which largely reduces the image quality. To address these issues, \citet{attnentropy} proposed to manually adjust the scale of entropy in the attention operations. 
\citet{scalecrafter} and \citet{fouriscale} attempted to enlarge the receptive field by replacing the original convolutional layers with strided ones, while \citet{hidiffusion} explicitly resizes the intermediate feature maps to match the training size.
%, which mitigated object duplications. 
\citet{demofusion} and \citet{accdiffusion} took a different strategy by generating a reference image at the base resolution and then using it to guide the whole sampling process at higher resolutions. 
Despite the great advancements, these methods still suffer from significant inference latency, hindering their broader applications in real world.

In this paper, %to expedite inference without compromising image quality, 
we propose an efficient \textbf{Fre}quency-aware \textbf{Ca}scaded \textbf{S}ampling framework, namely \textbf{FreCaS}, for training-free higher-resolution image generation. 
Our proposed FreCaS framework is based on the observation that latent diffusion models exhibit a coarse-to-fine generation manner in the frequency domain. In other words, they first generate low-frequency contents in early diffusion stages and gradually generate higher-frequency details in later stages. 
Leveraging this insight, we generate higher-resolution images through multiple stages of increased resolutions, progressively synthesizing details of increased frequencies.
FreCaS avoids unnecessary computations during the early diffusion stages as high-frequency details are not yet required.

In the latent space, the image representation expands its frequency range as the resolution increases. To encourage detail generation within the expanded frequency band, we introduce a novel frequency-aware classifier-free guidance (FA-CFG) strategy, which prioritizes newly introduced frequency components by assigning them higher guidance strengths in the sampling process.
Specifically, we decompose both unconditional and conditional denoising scores into two parts: low-frequency component, which captures content from earlier stages, and high-frequency component, which corresponds to the newly increased frequency band.
FA-CFG applies the classifier-free guidance to different frequency components with different strengths, and outputs the final denoising score by combining the adjusted components.
The FA-CFG strategy can synthesize much clear details while maintaining computational efficiency.
Additionally, to alleviate the issue of unfaithful layouts, such as duplicated objects mentioned in~\citet{attnentropy}, we reuse the cross-attention maps (CA-maps) from the previous stage, which helps maintaining consistency in image structure across different stages and ensuring more faithful object representations.

In summary, our main contributions are as follows:

\begin{itemize}[topsep=-5pt]
\item We propose FreCaS, an efficient frequency-aware cascaded sampling framework for training-free higher-resolution image generation. FreCaS leverages the coarse-to-fine nature of the latent diffusion process, thereby reducing unnecessary computations associated with  processing premature high-frequency details.
\item We design a novel FA-CFG strategy, which assigns different guidance strengths to components of different frequencies. This strategy enables FreCaS to focus on generating contents of  newly introduced frequencies in each stage, and hence synthesize clearer details. In addition, we fuse the CA-maps of previous stage and current stage to maintain a consistent image layouts across stages.
\item We demonstrate the efficiency and effectiveness of FreCaS through extensive experiments conducted on various pretrained diffusion models, including SD2.1, SDXL and SD3, validating its broad applicability and versatility.
\end{itemize}
\section{Related Works}

\subsection{Diffusion Models}
\label{m1}
Diffusion models have gained significant attentions due to their abilities to generate high-quality natural images.
\citet{ddpm} pioneered the use of a variance-preserving diffusion process to bridge the gap from natural images to pure noises.
\citet{adm} exploited various network architectures and achieved superior image quality than contemporaneous GAN models.
\citet{cfg} introduced a novel classifier-free guidance strategy that attains both generated image quality and diversity.
% without requiring training additional classifiers.
%
However, the substantial model complexity makes high-resolution image synthesis challenging.
\citet{cascadeddm} proposed a novel cascaded framework that progressively increases image resolutions.
\citet{ldm} performed the diffusion process in the latent space of a pre-trained autoencoder, enabling high-resolution image synthesis with reduced computational cost.
%
% On the other hand, some works~\citep{simplediffusion,importance,fdm,relay} have recognized that standard noise scheduling strategies are not optimized for higher resolutions. They propose adjusting the noise scheduling to maintain a constant signal-to-noise ratio (SNR) for intermediate states, resulting in better image quality.
%
\citep{sd3} presented SD3, which employs the rectified flow matching~\citep{rectifiedfm,rectifiedflow} at the latent space and demonstrates superior performance.
Despite the great progress, it still requires substantial efforts to train a high-resolution diffusion model from scratch.
Therefore, training-free higher-resolution image synthesis attracts increasing attentions.

\subsection{Training-free higher-resolution Image Synthesis}
\label{m3}

A few methods have been developed to leverage pre-trained diffusion models to generate images of higher resolutions than the training size.
MultiDiffusion~\citep{multidiffusion} is among the first methods to bind multiple diffusion processes into one unified framework and generates seamless higher-resolution images. 
However, the results exhibit unreasonable image structures such as duplicated objects.
% However, this method still faces the object duplication problem when applied to resolutions higher than the training resolution.
%
AttnEntropy~\citep{attnentropy} alleviates this problem by re-normalizing the entropy of attention blocks during sampling.
On the other hand, ScaleCrafter~\citep{scalecrafter} and FouriScale~\citep{fouriscale} expand the receptive fields of pre-trained networks to match higher inference resolutions, thereby demonstrating improved image quality.
HiDiffusion~\citep{hidiffusion} dynamically adjusts the feature sizes to match the training dimensions.
DemoDiffusion~\citep{demofusion} and AccDiffusion~\citep{accdiffusion} first generate a reference image at standard resolutions and then use this image to guide the generation of images at higher resolutions.
Despite their success, the above mentioned approaches neglect the coarse-to-fine nature of image generation and generate image contents of all frequencies simultaneously, resulting in long inference latency and limiting their broader applications.

To address this issue, we propose an efficient FreCaS framework for training-free higher-resolution image synthesis. FreCaS divides the entire sampling process into stages of increasing resolutions, gradually synthesizing components of different frequency bands, thereby reducing the unnecessary computation of handling premature high-frequency details in early sampling stages.
\rebuttal{It is worth noting that DemoFusion~\citep{demofusion} and ResMaster~\citep{resmaster} also employ a cascaded sampling scheme. However, there exist fundamental differences between FreCaS and them: DemoFusion and ResMaster perform a complete diffusion process at each resolution, whereas FreCaS transitions the diffusion from low to high resolutions in just one process. 
This distinction makes our method significantly more efficient than them while achieving better image quality.}
%
%It is worth noting that DemoFusion~\citep{demofusion} employs a progressive upscaling scheme that may appear similar to our approach. However, there exist fundamental differences: DemoFusion utilizes the generated image at the base resolution as a reference to guide the entire sampling process at target resolution, whereas FreCaS continues the sampling process at higher resolutions. 

\section{Method}
This section presents the details of the proposed FreCaS framework,
%, which aims to achieve a fast higher-resolution image synthesis framework using pretrained latent diffusion models.
%
which leverages the coarse-to-fine nature of latent diffusion models and constructs a frequency-aware cascaded sampling strategy to progressively refine high-frequency details.
We first introduce the notations and concepts that form the basis of our approach (see Section~\ref{m1}).
Then, we delve into the key components of our method: FreCaS framework (see Section~\ref{m2}), FA-CFG strategy (see Section~\ref{m3}), and CA-maps re-utilization (see Section~\ref{m4}). 
%Each of these components is designed to work synergistically to enhance the overall performance of our approach.

\subsection{Preliminaries}
\label{m1}
\textbf{Diffusion models.}
Diffusion models~\citep{ddpm,adm} transform complex image distributions into the Gaussian distribution, and vice versa. They gradually inject Gaussian noises into the image samples, and then use a reverse process to remove noises from them, achieving image generation.
%
% Some popular applications of diffusion models include SD2.1, SDXL and SD3.
%
% The first two models, SD2.1 and SDXL, are build upon the classical latent diffusion framework, while SD3 switches to a more advanced diffusion framework, rectified flow matching, which straightens the diffusion trajectories to achieve fast convergence.
%
Most recent diffusion models operate in the latent space and utilize a discrete timestep sampling process to synthesize images.
Specifically, for a $T$-step sampling process, a latent noise \(\vz_T\) is drawn from a standard Gaussian distribution, and then iteratively refined through a few denoising steps until converged to the clean signal latent \(\vz_0\). Finally, the natural image \(\vx\) is decoded from \(\vz_0\) using a decoder $\mathcal{D}$. The whole process can be written as follows:
\begin{equation}
    \vz_T \sim \mathcal{N}(\textbf{0}, \textbf{I}) \rightarrow \vz_{T-1} \rightarrow \cdots \rightarrow \vz_{\text{1}} \rightarrow \vz_{\text{0}} \rightarrow \vx = \mathcal{D}(\vz_{\text{0}}).
\end{equation}
% For each denoising step, current works predict denoising scores $\hat \bepsilon$ using a network and sample the previous state $\vz_{t-1}$ from $\mathcal{N}(\vmu(\vz_t, \hat \vz_\text{0}(\vz_t, \hat \bepsilon)), \bsigma^2)$,
% \begin{equation}
% 	\vz_{t-1} \sim \mathcal{N}(\vmu(\vz_t, \hat \vz_\text{0}), \bsigma^2) = \mathcal{N}(\vmu(\vz_t, \frac {\vz_t - \sqrt {1-\alpha_t}\hat {\bepsilon}(\vz_t)}{\sqrt {\alpha_t}}), \bsigma^2),
% \end{equation}
% where $\alpha_1,\cdots, \alpha_\text{T}$ denote the noise schedule of diffusion processes. $\mu$ and $\sigma$ are the mean and standard deviation of the posterior sampling distribution, respectively.
% A few works~\citep{} further proposal to forecast the variance $\sigmas$ during sampling, showing some improvements.

% Additionally, techniques such as classifier-free guidance (CFG)~\citep{cfg} can be further employed to improve image quality.
For each denoising step, current works typically adopt the classifier-free guidance (CFG)~\citep{cfg} to improve image quality.
It predicts an unconditional denoising score $\bepsilon_{unc}$ and a conditional denoising score $\bepsilon_{{c}}$.
The final denoising score is obtained via a simple extra-interpolation process as $\hat{\bepsilon} = (1-w) \cdot \bepsilon_{unc} + w \cdot \bepsilon_{c}$, where $w$ denotes the guidance strength. 
% \begin{equation}
% \hat{\bepsilon} = (1-w) \cdot \bepsilon_{unc} + w \cdot \bepsilon_{c},
% \end{equation}

% When \(w\) is large, the strategy emphasizes generating text-aligned objects and clean details. Conversely, when \(w\) is small, CFG tends to synthesize images more naturally.

\textbf{Resolution and frequency range.} The resolution of a latent $\vz$ determines its sampling frequency~\citep{heat}, thereby influencing its frequency domain characteristics.
Specifically, if a latent of unit length has a resolution of $s \times s$, its sampling frequency $f_s$ can be defined as the number of samples per unit length, which is $s$.
The Nyquist frequency is then obtained as $\frac{f_s}{2} = \frac{s}{2}$. Therefore, the frequency of the latent $\vz$ ranges from [0, $\frac s{2}$]. 
Reducing its resolution to $s_l\times s_l$ narrows the frequency range to [0, $\frac{s_l}{2}$].
As a result, higher resolutions capture a broader frequency domain, while lower resolutions lead to a narrower frequency spectrum. 
%This relationship is crucial for understanding how resolution impacts the frequency characteristics of latents.
\begin{figure*}[!t]
%\vspace{10pt}
%\begin{overpic}[width=\linewidth]{figures/statistics.jpg}
\begin{overpic}[width=\linewidth]{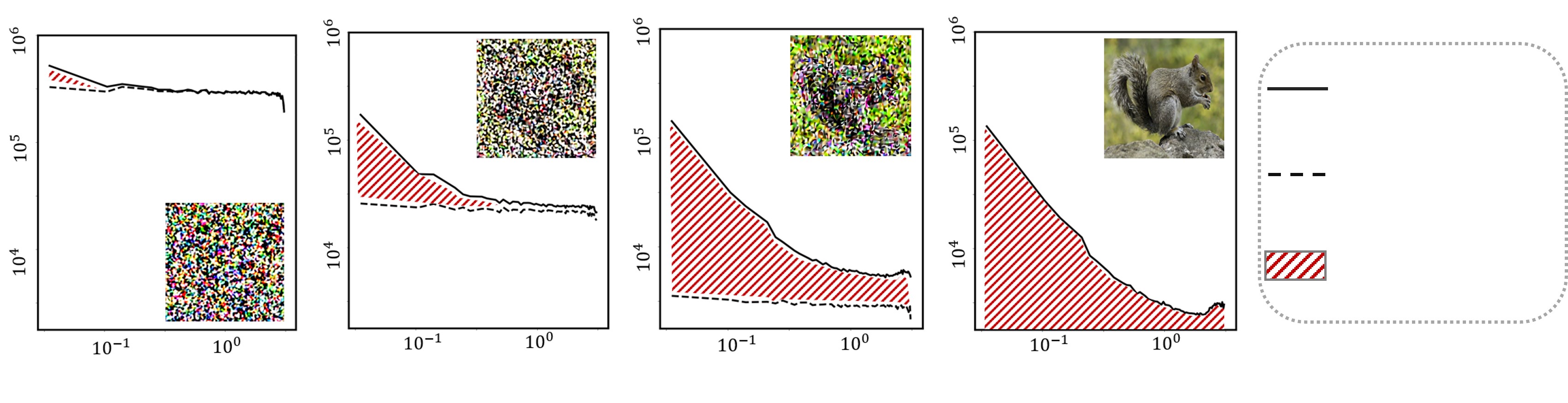}
\put(-2,5){\rotatebox{90}{\scriptsize{power spectral density (PSD)}}}
\put(-1.5,3){\tiny{Low Freq.}}
\put(75.5,2.5){\tiny{High Freq.}}
\put(6,0){\footnotesize{(a) $\vz_{900}$}}
\put(27,0){\footnotesize{(b) $\vz_{600}$}}
\put(47,0){\footnotesize{(c) $\vz_{300}$}}
\put(68,0){\footnotesize{(d) $\vz_{0}$}}
\put(85.3,19.5){\scriptsize{PSD of latents}}
\put(85.3,14){\scriptsize{PSD of noise parts}}
\put(85.3,8){\scriptsize{PSD of signal parts}}
\end{overpic}
\vspace{-10pt}
\caption{From (a) to (d), the sub-figures show the PSD curves of latents $\vz_\text{900}$, $\vz_\text{600}$, $\vz_\text{300}$ and $\vz_\text{0}$ of SDXL, respectively.
%curves of $z_0$, curves of $z_{300}$, curves of $z_{600}$ and curves of $z_{900}$. 
One can see that the energy of synthesized clean signals (the red slashed regions) first emerges in the low-frequency band and gradually expands to high-frequency band.
% Solid lines denotes the PSD curves of cooresponding latents while dashed lines represents the PSD curve of noisy part of that variables.}
}

\label{fig:fig_statistics}
\end{figure*}

\subsection{Frequency-aware cascaded sampling}
\label{m2}

Pixel space diffusion models exhibit a coarse-to-fine behavior in the image synthesis process~\citep{heat,relay}. 
% Those methods sequentially produces content of varying frequency components during the sampling process, ensuring a gradual refinement of image details. 
In this section, we show that such a behavior is also exhibited for latent diffusion models during the sampling process, which inspires us to develop a frequency-aware cascaded sampling framework for generating higher-resolution images.

\textbf{PSD curves in latent space.} The power spectral density (PSD) is a powerful tool for analyzing the energy distribution of signals along the frequency spectrum. \citet{heat} and \citet{relay} have utilized PSD to study the behaviour of intermediate states in the pixel diffusion process.
% Following the guidelines of \citet{heat,relay} in pixel space, we also draw the PSD curves in the latent space to observe the behavior of diffusion processes from a frequency perspective.
%
Here, we compute the PSD of the latent signals over a collection of 100 natural images using the pre-trained SDXL model~\citep{sdxl}.
Figure~\ref{fig:fig_statistics} shows the PSD curves of $\vz_\text{900}$, $\vz_\text{600}$, $\vz_\text{300}$ and $\vz_\text{0}$.
The solid line denotes the PSD curve of intermediate noise corrupted latent, while the dashed line represents the PSD of Gaussian noise corrupted into the latent.
The inner area between the two curves (marked with red slashes) indicates the energy of clean signal latent being synthesized.
% Figure~\ref{fig:fig_statistics} presents the PSD curves for the latents of SDXL and SD3 across a collection of 100 natural images. 
% Specifically, from left to the right, the four subfigures show the PSD curves of $\vz_{\text{0}}$, $\vz_\text{300}$, $\vz_\text{600}$, and $\vz_\text{900}$, illustrating how they change throughout the diffusing process.
%
% It is evident from the figure that the PSD of the clear latents $\vz_\text{0}$ (the solid black curve in the most left subfigure) follows a power-law distribution of $1/f^{\alpha}$, similar to the spectral characteristics of natural images in pixel space~\citep{modelling,statistics,heat}. 
%
% The PSD curves for all the noise components (dashed lines) exhibit uniform values across all frequency bands.
%
% On the other hand, as the sampling process progresses, the PSD curves of the latents and the noise first separate in the low-frequency range, as exemplified by the PSD curves of $\vz_\text{900}$ (green lines), indicating the generation of low-frequency content. 
%
% This is followed by a separation in the high-frequency range, as illustrated by the PSD curves of $\vz_\text{600}$ (red lines) and $\vz_\text{300}$ (blue lines), corresponding to the synthesis of high-frequency details. 
%
One can see that the clean image signals emerge from  the low-frequency band (see $\vz_\text{900}$ and $\vz_\text{600}$) and gradually expand to the high-frequency band (see $\vz_\text{300}$ and $\vz_\text{0}$) during the sampling process.
These observations confirm the coarse-to-fine nature of image synthesis in the latent diffusion process, where low-frequency content is generated first, followed by high-frequency details.

\begin{figure*}[!t]
%\vspace{10pt}

%\begin{overpic}[width=\linewidth]{figures/framework.jpg}
\begin{overpic}[width=\linewidth]{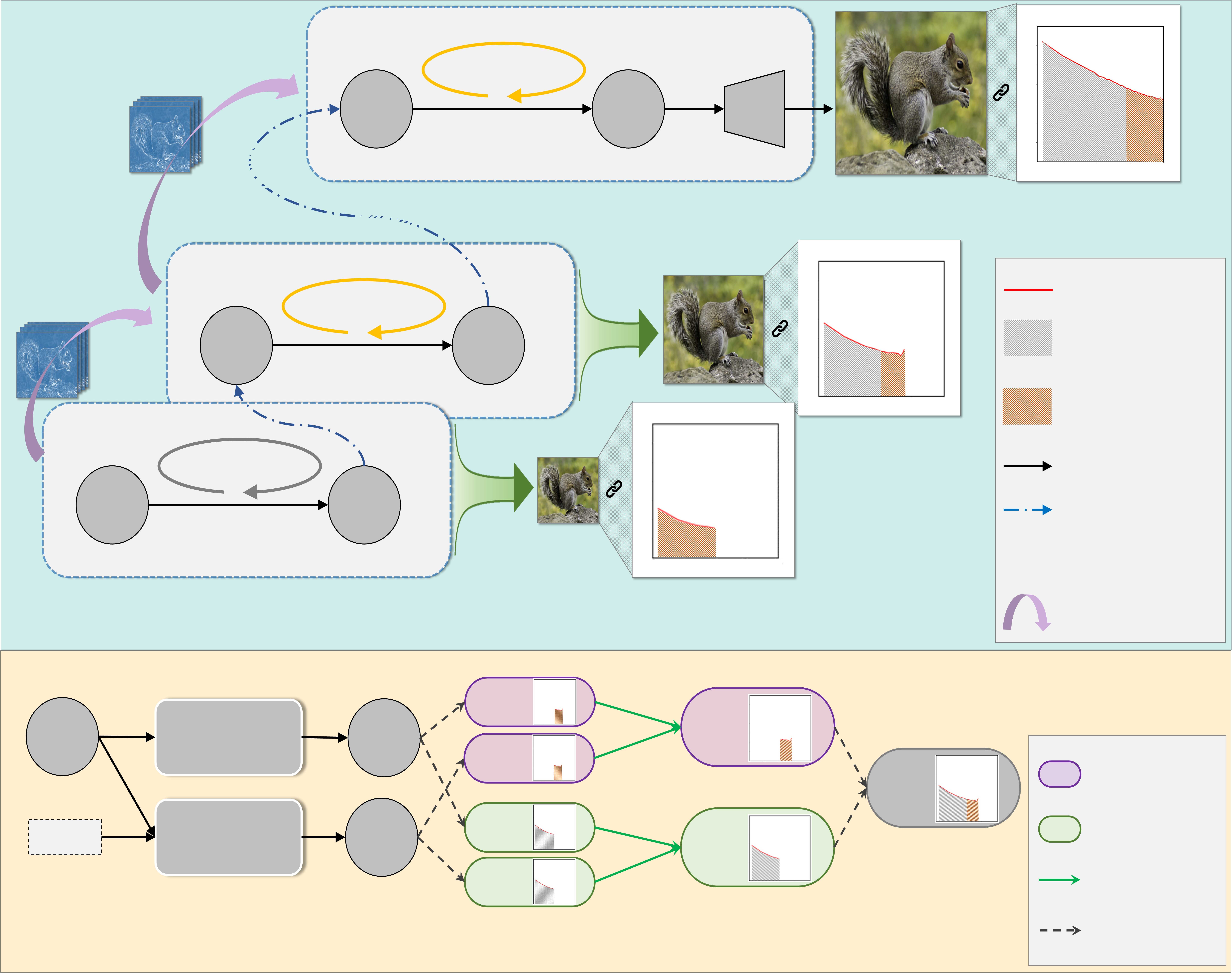}
\put(1,49.5){\scriptsize{CA-maps}}
\put(10,66.5){\scriptsize{CA-maps}}

\put(4.5,40){\scriptsize{\text{Stage} \text{0}}}
\put(7.5,34.5){\footnotesize{$\vz^{s_0}_T$}}
\put(28.2,34.5){\footnotesize{$\vz^{s_0}_L$}}
\put(17.5,37.3){\scriptsize{CFG}}

\put(14,52){\scriptsize{\text{Stage} \text{1}}}
\put(17.5,46.5){\footnotesize{$\vz^{s_1}_F$}}
\put(38.2,46.5){\footnotesize{$\vz^{s_1}_L$}}
\put(26.8,49.3){\scriptsize{FA-CFG}}

\put(22.5,57.3){\scriptsize{\text{Stage} \textbf{$i$}}}
\put(26,70){\scriptsize{\text{Stage} \textbf{$N$}}}
\put(28.8,64.2){\footnotesize{$\vz^{s_N}_F$}}
\put(49.6,64.2){\footnotesize{$\vz^{s_N}_0$}}
\put(38,67){\scriptsize{FA-CFG}}
\put(60.5,64){\footnotesize{D}}

\put(85,71.4){\scalebox{0.8}{\tiny{PSD along Freq.}}}
\put(67.5,53.8){\scalebox{0.8}{\tiny{PSD along Freq.}}}
\put(54,41.6){\scalebox{0.8}{\tiny{PSD along Freq.}}}
\put(35,26){\footnotesize{(a) FreCaS Framework}}

\put(86.3,50.5){\scriptsize{PSD curves}}
\put(86.3,47){\scriptsize{\makecell[l]{Freq. bands of \\previous stages}}}
\put(86.3,42){\scriptsize{\makecell[l]{Expanded Freq. \\bands}}}
\put(86.3,37.5){\scriptsize{Sample process}}
\put(86.3,34.3){\scriptsize{Transition process}}
\put(81,30.4){\scalebox{0.75}{\scriptsize{$\vz^{s_i}_F$, $\vz^{s_i}_L$}}}
\put(86.3,30.4){\scriptsize{\makecell[l]{The first/last state\\ of stage $i$}}}
\put(86.3,26.4){\scriptsize{CA-maps reuse}}

\put(4,17){\footnotesize{$\vz_t$}}
\put(3.5,9.5){\footnotesize{text}}
\put(14,17){\scriptsize{Uncond Net}}
\put(15,9.5){\scriptsize{Cond Net}}
\put(29,17){\footnotesize{$\bepsilon_{unc}$}}
\put(30,9.5){\footnotesize{$\bepsilon_{c}$}}
\put(39,19.8){\scriptsize{$\bepsilon^h_{unc}$}}
\put(40,15.5){\scriptsize{$\bepsilon^h_{c}$}}
\put(39,10.3){\scriptsize{$\bepsilon^l_{unc}$}}
\put(40,6){\scriptsize{$\bepsilon^l_{c}$}}
\put(49.5,20.5){\rotatebox{-15} {\scriptsize{1-$w_h$}}}
\put(49.5,17.3){\rotatebox{15} {\scriptsize{$w_h$}}}
\put(49.5,11.3){\rotatebox{-12} {\scriptsize{1-$w_l$}}}
\put(49.5,8){\rotatebox{15} {\scriptsize{$w_l$}}}
\put(57.5,17.5){\footnotesize{$\hat \bepsilon^h$}}
\put(57.5,8.5){\footnotesize{$\hat \bepsilon^l$}}
\put(73,13){\footnotesize{$\hat \bepsilon$}}

\put(88.3,14.3){\scriptsize{High Freq. part}}
\put(88.3,10.15){\scriptsize{Low Freq. part}}
\put(88.3,6.5){\scriptsize{CFG}}
\put(88.3,2.8){\tiny{\makecell[l]{Freq. decompose\\or Freq. combine}}}
\put(41,1.5){\footnotesize{(b) FA-CFG}}
\end{overpic}

\vspace{-5pt}

\caption{
	(a) The overall framework of FreCaS. The entire $T$-step sampling process is divided into $N+1$ stages of increasing resolutions and expanding frequency bands. FreCaS starts the sampling process at the training size and obtains the last latent $\vz^{s_\text{0}}_{L}$ at that stage. Then, FreCaS continues the sampling from the first latent $\vz^{s_\text{1}}_F$ at the next stage with a larger resolution and expanded frequency domain. This procedure is repeated until the final latent $\vz^{s_N}_\text{0}$ at stage $N$ is obtained. A decoder is then used to generate the final image.
	(b) FA-CFG strategy. 
	% FreCaS leverages the FA-CFG strategy to direct diffusion models more on generating contents of expanded frequency bands. 
	We separate the original denoising scores into low-frequency and high-frequency components and assign a higher CFG strength to the high-frequency part. The two parts are then combined to obtain the final denoising score $\hat \bepsilon$.
	% The top subfigure shows the overall pipeline of our FreCaS framework while the bottom part presents the details of FA-CFG strategy. Please zoom-in for better details.
	}
\label{fig:fig_framework}

\vspace{-15pt}
\end{figure*}

\textbf{Framework of FreCaS.} Based on the above observation, we developed an efficient FreCaS framework to progressively generate image contents of higher frequency bands, reducing unnecessary computations in processing premature high-frequency details in early diffusion stages.
As shown in Figure~\ref{fig:fig_framework}(a), our FreCaS divides the entire $T$-step sampling process into $N+1$ stages of increasing resolutions.
The initial stage performs the sampling process at the default training size $s_\text{0}$ with a frequency range of [0, $\frac {s_\text{0}}2$].
Each of the subsequent stages increases the sampling size to its predecessor, gradually expanding the frequency domain.
At the final stage, the latent reaches the target resolution $s_{N}$, achieving a full frequency range from 0 to $\frac {s_{N}}{2}$. 

Specifically, we begin with a pure noise latent \( \vz^{s_\text{0}}_{T} \) at stage $ s_\text{0} $, and iteratively perform reverse sampling until obtaining the last latent in this stage, denoted by \( \vz^{s_\text{0}}_{L} \).
Next, we transition \( \vz^{s_\text{0}}_{L} \) to the first latent, denoted by \( \vz^{s_\text{1}}_{F} \),  in next stage, as illustrated by the blue dashed arrow in Figure~\ref{fig:fig_framework}(a). 
This procedure is repeated until the latent feature reaches the target size, resulting in \( \vz^{s_{N}}_{\text{0}} \). The final image \( \vx \) is obtained by applying the decoder to \( \vz^{s_{N}}_{\text{0}} \) so that \( \vx = \mathcal{D}(\vz^{s_{N}}_{\text{0}}) \). 
With such a sampling pipeline, FreCaS ensures a gradual refinement of details across coarse-to-fine scales, ultimately producing a high-quality and high-resolution image with minimum computations.

\label{transition_process}
For the transition between two adjacent stages, we perform five steps to convert the last latent of previous stage $\vz^{s_{i-1}}_{L}$ to the first latent of next stage $\vz^{s_i}_{F}$:
\begin{equation}
    \vz^{s_{i-1}}_{L} \xrightarrow{\text{denoise}} \hat \vz^{s_{i-1}}_\text{0} \xrightarrow{\text{decode}} \hat \vx^{s_{i-1}} \xrightarrow{\text{interpolate}} \hat \vx^{s_{i}} \xrightarrow{\text{encode}} \vz^{s_i}_{0} \xrightarrow{\text{diffuse}} \vz^{s_i}_{F},
\end{equation}
where ``denoise" and ``diffuse" are standard diffusion operations, ``decode" and ``encode" are performed using the decoder and encoder, respectively, and ``interpolation" adjusts the resolutions using the bilinear interpolation. 
To determine the timestep of $\vz^{s_i}_{F}$, we follow previous works~\citep{simplediffusion,importance,fdm,relay} to keep the signal-to-noise ratio (SNR) equivalence between \( \vz_{{L}}^{s_{i-1}} \) and \( \vz_{F}^{s_{i}} \).
% Please refer to Appendix~\ref{appendix_transition} for more details.
Please refer to Appendix \appendixid{shifting_timestep} for more details.

\subsection{FA-CFG strategy}
\label{m3}

Our FreCaS framework progressively transitions the latents to stages with higher resolutions and extended high-frequency bands. To ensure that the diffusion models focus more on generating contents of newly introduced frequencies, we propose a novel FA-CFG strategy, which assigns higher guidance strength to the new frequency components.
In FreCaS, upon transitioning to stage $s_i$, the latent increases its resolution from $s_{i-1}$ to the higher resolution $s_i$, thereby expanding the frequency band from [0, $\frac{s_{i-1}}{2}$] to [0, $\frac{s_{i}}{2}$].
This inspires us to divide the latents into two components: a low-frequency component ranging from [0, $\frac{s_{i-1}}{2}$] and a high-frequency component covering the frequency interval ($\frac{s_{i-1}}{2}$, $\frac{s_i}{2}$].
The former preserves the generated contents from previous stages, whereas the latter is reserved for the contents to be generated in this stage.
Our goal is to encourage the diffusion models to generate natural details and textures in the newly expanded frequency band.

To achieve the above mentioned goal, we propose to perform CFG on the two frequency-aware parts with different guidance strengths.
The entire process is illustrated in Figure~\ref{fig:fig_framework}(b).
First, we obtain the unconditional denoising score $\mathbf{\bepsilon}_{unc}$ and conditional denoising score $\mathbf{\bepsilon}_{c}$ using the pre-trained diffusion network. 
Then, we split the scores into a low-frequency part and a high-frequency part.
The former is extracted by downsampling the scores and then resizing them back, while the latter is the residual by subtracting the low-frequency part from the original denoising scores.
Subsequently, we apply the CFG strategy to the two parts with different weights. 
Specifically, for the low-frequency part, we assign the normal guidance strength $w_l$, while for the high-frequency part, we use a much higher weight $w_h$ to prioritize content generation in this frequency band.
The final denoising score is obtained by summing up the two parts. This process can be expressed as:
\begin{equation}
\hat{\bepsilon} = \hat \bepsilon^l + \hat \bepsilon^h = (1-w_l)\cdot \bepsilon^l_{unc} + w_l\cdot \bepsilon^l_{c} + (1-w_h)\cdot \bepsilon^h_{unc} + w_h\cdot \bepsilon^h_{c},
\end{equation}
where $\hat \bepsilon^l$ and $\hat \bepsilon^h$ are the low-frequency and high-frequency parts of $\hat \bepsilon$, respectively. Similarly, $\bepsilon^l_{unc}$, $\bepsilon^h_{unc}$, $\bepsilon^l_{c}$ and $\bepsilon^h_{c}$ follow the same notation.

\subsection{CA-maps reutilization}
\label{m4}
When applied to higher resolutions, pre-trained diffusion models often present unreasonable image structures, such as duplicated objects.
To address this issue, we propose to reuse the CA-maps from the previous stage to maintain layout consistency across stages. 
The CA-maps represent attention weights from cross-attention interactions between spatial features and textual embeddings, effectively capturing the semantic layout of the generated images.
Specifically, we average the CA-maps of all cross-attention blocks when predicting $\vz^{s_{i-1}}_{L}$ at stage $s_{i-1}$.
After transitioning to stage $s_i$, we replace the current CA-maps of each cross-attention block using its linear interpolation with the averaged CA-maps $\overline M^{s_{i-1}}_{{L}}$ as follows:
\begin{equation}
    M^{s_{i}}_t = (1-w_c) \cdot M^{s_{i}}_t + w_c \cdot \overline M^{s_{i-1}}_{{L}},
\end{equation}
where \(M^{s_{i}}_t\) is the CA-maps at step \(t\) of  stage \(s_{i}\). 
In this way, FreCaS can effectively maintain content consistency and prevent unexpected objects or textures during higher-resolution image generation.

\vspace{-5pt}
\section{Experiments}

\begin{table*}[!t]

\centering
\renewcommand{\arraystretch}{1.2}

\vspace{-10pt}

\caption{\rebuttal{Experiments on $\times 4$ and $\times 16$ generation of SD2.1 and SDXL. ``DO" means ``duplicated object", which indicates whether the method takes the duplicated object problem into consideration. ``SpeedUP" denotes the efficiency speed-up over the DirectInference baseline. The \best{red} and \second{blue} indicate the best and second ones among all methods that consider the duplicated object problem.\\}}

\vspace{-5pt}

\label{tab:sd21_sdxl}

\scalebox{0.93}{
\begin{tabular}{>{\centering\arraybackslash}p{5pt}|>{\centering\arraybackslash}p{5pt}|c|>{\centering\arraybackslash}p{11pt}|>{\centering\arraybackslash}p{19pt}>{\centering\arraybackslash}p{19pt}>{\centering\arraybackslash}p{19pt}>{\centering\arraybackslash}p{19pt}>{\centering\arraybackslash}p{19pt}>{\centering\arraybackslash}p{30pt}|>{\centering\arraybackslash}p{36pt}>{\centering\arraybackslash}p{30pt}}
    \hline
%    \rotatebox{90}{Scales} & \rotatebox{90}{Models\;} & \makecell[c]{Methods} & \rotatebox{90}{\makecell[c]{\small{Duplicated\blank}\\ \scriptsize{Objects}}} & \footnotesize{FID$\downarrow$} & \footnotesize{$\text{FID}_p$$\downarrow$} & \footnotesize{IS$\uparrow$} & \footnotesize{$\text{IS}_p$$\uparrow$} &  \makecell[c]{\scriptsize{CLIP}\\ \scriptsize{SCORE}$\uparrow$} & \footnotesize{Latency(s)$\downarrow$} & \footnotesize{SpeedUP$\uparrow$} \\
\diagbox{}{} & \diagbox{}{} & Methods & DO & \footnotesize{FID} & \footnotesize{$\text{FID}_b$$\downarrow$} & \footnotesize{$\text{FID}_p$$\downarrow$} & \footnotesize{IS$\uparrow$} & \footnotesize{$\text{IS}_p$$\uparrow$} &  \makecell[c]{\scriptsize{CLIP}\\[-0.3em] \scriptsize{SCORE}}$\uparrow$ & \footnotesize{Latency(s)$\downarrow$} & \footnotesize{SpeedUP$\uparrow$} \\
    \hline
    \hline
    
\multirow{14}*{\rotatebox{90}{SD2.1}} & \multirow{7}*{\rotatebox{90}{$\times$4}} & DirectInference & \xmark & 31.07 & 34.54 & 23.84 & 15.00 & 17.26 & 32.01 & 5.50 & 1x \\
& & MultiDiffusion & \xmark & 21.05  & 22.44 & 14.68  & 17.46 & 18.29 & 32.49 & 120.21 & 0.046$\times$\\
\cdashline{3-12}
& & AttnEntropy & \cmark & 28.33 & 30.63 & \second{21.34} & 15.67 & \best{17.71} & 32.28 & 5.56 & 0.99$\times$ \\
& & ScaleCrafter & \cmark & \second{16.65} & \second{13.18} & 22.44 & \second{17.42} & \second{16.29} & \second{32.88} & 6.36 & 0.86$\times$\\
& & FouriScale & \cmark & 19.01 & 15.33 & 23.26 & 17.11 & 15.57 & \best{32.92} & 11.06 & 0.50$\times$ \\
& & HiDiffusion & \cmark & 19.95 & 16.21 & 25.26 & 17.13 & 16.12 & 32.37 & \second{3.57} & \second{1.54$\times$}\\
& & \colorbox{gray!30}{\textbf{\qquad Ours\qquad}} & \cmark & \best{16.38} & \best{13.14} & \best{21.23} & \best{17.55} & 16.04 & 32.33 & \best{2.56} & \best{2.16$\times$}\\
\cline{2-12}

 &  \multirow{7}*{\rotatebox{90}{$\times$16}}  & DirectInference & \xmark & 124.5 & 128.3 & 50.23 & 8.84 & 15.30 & 27.67 & 49.27 & 1$\times$ \\
& & MultiDiffusion & \xmark & 67.44 & 74.15 & 15.28 & 8.75 & 18.82 & 31.14 & 926.33 & 0.05$\times$ \\
\cdashline{3-12}
& & AttnEntropy & \cmark &122.6 &  127.6 & \second{46.52} & 9.31 & \best{16.25} & 28.33 & 49.33 & 1.00$\times$\\
& & ScaleCrafter & \cmark & 34.47 & 34.55 & 57.47 & 13.02 & 12.12 & 31.44 & 92.86 & 0.53$\times$\\
& & FouriScale & \cmark & 34.17& \second{34.13} & 58.01 & 12.79 & 13.15 & \second{31.68} & 90.13 & 0.55$\times$\\
& & HiDiffusion & \cmark & \second{33.15} & {34.17} & 70.58 & \second{13.49} & 11.87 & 31.09 & \second{18.22} & \second{2.70$\times$} \\
& & \colorbox{gray!30}{\textbf{\qquad Ours\qquad}} & \cmark & \best{19.95} & \best{20.11} & \best{43.71} & \best{15.22} & \second{13.74} & \best{31.92} & \best{13.35} & \best{3.69$\times$}\\
\hline

\multirow{16}*{\rotatebox{90}{SDXL}} &  \multirow{8}*{\rotatebox{90}{$\times$4}}  & DirectInference & \xmark & 39.15 & 43.83 & 29.71 & 11.52 & 14.60 & 32.51 & 34.10 & 1$\times$ \\
    \cdashline{3-12}
    & & AttnEntropy & \cmark & 36.54 & 41.30 & 27.67 & 11.69 & 15.04 & 32.71 & 34.36 & 0.99$\times$ \\
    & & ScaleCrafter & \cmark & 22.76 & 24.23 & 23.17 & 14.10 & 14.97 & 32.70 & 39.64 & 0.86$\times$ \\
    & & FouriScale & \cmark &26.44& 26.88 & 27.24 & 13.97 & 14.44 & 32.90 & 66.18 & 0.52$\times$\\
    & & HiDiffusion & \cmark &21.67& 20.69 & 21.80 & 15.56 & 15.93 & 32.62 & \second{18.38} & \second{1.86$\times$}\\
    & & AccDiffusion  & \cmark &19.87& 17.62 & 21.11 & 17.07 & 16.15 & 32.66 & 102.46 & 0.33$\times$\\ 
    & & DemoFusion & \cmark &\second{18.77}& \second{16.33} & \second{18.77} & \second{17.10} & \second{17.21} & \second{33.16} & 83.95 & 0.41$\times$\\
    & & \colorbox{gray!30}{\textbf{\qquad Ours\qquad}} & \cmark &\best{16.48}& \best{12.63} & \best{17.91} & \best{17.18} & \best{17.31} & \best{33.28} & \best{13.84} & \best{2.46$\times$} \\
    \cline{2-12}

    &  \multirow{8}*{\rotatebox{90}{$\times$16}}  & DirectInference & \xmark &145.4&  151.3 & 62.39 & 6.41 & 11.66 & 28.24 & 312.36 & 1$\times$ \\
    \cdashline{3-12}
    & & AttnEntropy & \cmark & 142.1& 148.9 & 60.54 & 6.46 & 12.44 & 28.46 & 312.46 & 1.00$\times$ \\
    & & ScaleCrafter & \cmark &71.49& 75.11 & 73.21 & 8.68 & 9.81 & 30.76 & 560.91 & 0.56$\times$ \\
    & & FouriScale &\cmark &98.01&  77.63 & 84.05 & 8.00 & 9.41 & 30.78 & 534.08 & 0.58$\times$  \\
    & & HiDiffusion & \cmark &81.48& 83.41 & 120.1 & 9.79 & 9.56 & 29.18 & \second{101.59} & \second{3.07$\times$}\\
    & & AccDiffusion  & \cmark &50.47& 48.15 & 46.07 & 12.11 & 11.75 & 32.26 & 763.23 & 0.41$\times$ \\
    & & DemoFusion & \cmark &\second{47.80}& \second{44.54} & \best{35.52} & \second{12.38} & \second{13.82} & \best{33.03} & 649.25 & 0.48$\times$ \\
    & & \colorbox{gray!30}{\textbf{\qquad Ours\qquad}} & \cmark &\best{42.75}& \best{40.63} & \second{39.82} & \best{12.68} & \best{14.16} & \best{33.03} & \best{85.87} & \best{3.64$\times$} \\

% & & \colorbox{gray!30}{\textbf{\qquad Ours\qquad}} & \cmark & \best{40.63} & \second{39.31} & \second{12.21} & \best{14.19} & \best{33.03} & \best{85.87} & \best{3.64$\times$} \\
    \hline
    \hline

\end{tabular}
}

\vspace{-20pt}
\end{table*}

\begin{figure*}[!t]

\vspace{-1pt}

%\begin{overpic}[width=0.97\linewidth]{figures/sd21.p1.jpg}
% \begin{overpic}[width=0.97\linewidth]{figures/sd21.notext.jpg}
\begin{overpic}[width=\linewidth]{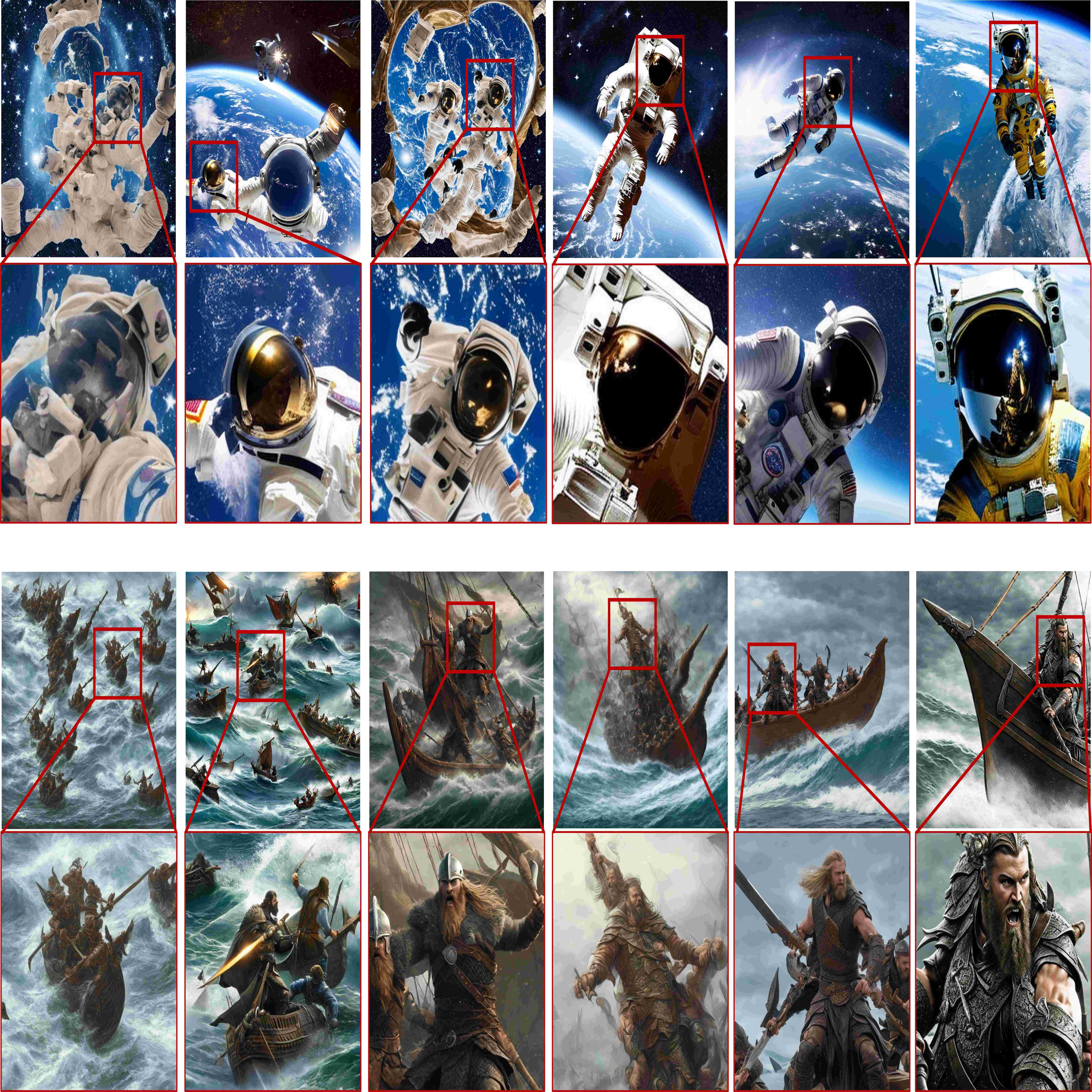}
\put(-3,46){\rotatebox{90}{\footnotesize {$\times 4$ on SD2.1}}}
\put(-3,10){\rotatebox{90}{\footnotesize {$\times 16$ on SD2.1}}}

\put(2,33.5){\scriptsize{DirectInference}}
\put(19,33.5){\scriptsize{MultiDiffusion}}
\put(37.5,33.5){\scriptsize{AttnEntropy}}
\put(54.5,33.5){\scriptsize{ScaleCrafter}}
\put(71,33.5){\scriptsize{HiDiffusion}}
\put(87.5,33.5){\scriptsize{FreCaS(\textbf{ours})}}

\put(2,-1.8){\scriptsize{DirectInference}}
\put(19,-1.8){\scriptsize{MultiDiffusion}}
\put(37.5,-1.8){\scriptsize{AttnEntropy}}
\put(54.5,-1.8){\scriptsize{ScaleCrafter}}
\put(71,-1.8){\scriptsize{HiDiffusion}}
\put(87.5,-1.8){\scriptsize{FreCaS(\textbf{ours})}}
\end{overpic} 

\vspace{15pt}

% \begin{overpic}[width=0.97\linewidth]{figures/sdxl.p1.jpg}
% \begin{overpic}[width=0.97\linewidth]{figures/sdxl.notext.jpg}
\begin{overpic}[width=\linewidth]{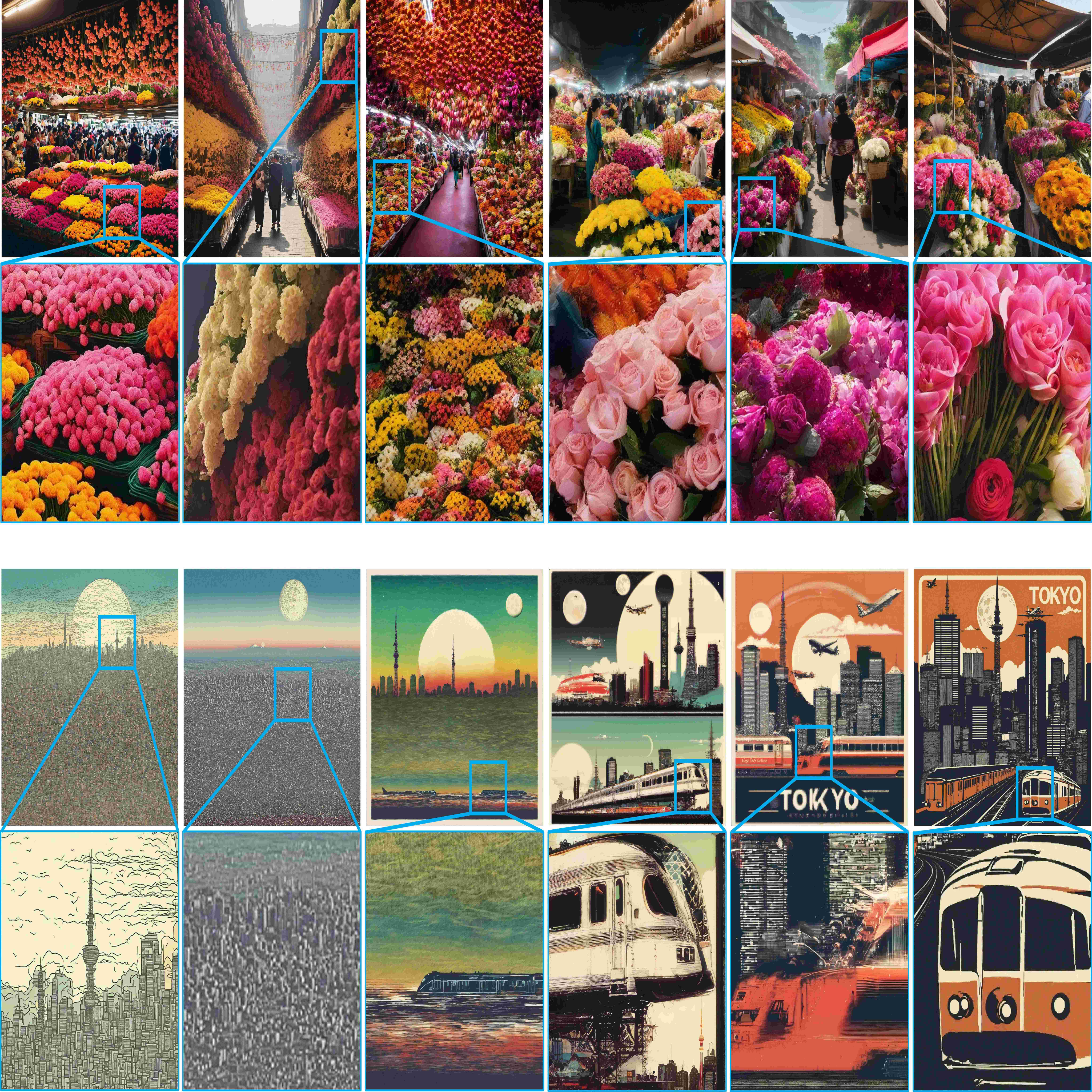}
\put(-3,46){\rotatebox{90}{\footnotesize {$\times 4$ on SDXL}}}
\put(-3,10){\rotatebox{90}{\footnotesize {$\times 16$ on SDXL}}}

\put(3,34){\scriptsize{ScaleCrafter}}
\put(21,34){\scriptsize{FouriScale}}
\put(37.5,34){\scriptsize{HiDiffusion}}
\put(53.5,34){\scriptsize{AccDiffusion}}
\put(70.5,34){\scriptsize{DemoFusion}}
\put(87.5,34){\scriptsize{FreCaS(\textbf{ours})}}

\put(3,-1.8){\scriptsize{ScaleCrafter}}
\put(21,-1.8){\scriptsize{FouriScale}}
\put(37.5,-1.8){\scriptsize{HiDiffusion}}
\put(53.5,-1.8){\scriptsize{AccDiffusion}}
\put(70.5,-1.8){\scriptsize{DemoFusion}}
\put(87.5,-1.8){\scriptsize{FreCaS(\textbf{ours})}}
\end{overpic}

% \vspace{0pt}

\caption{Visual comparison on $\times 4$ and $\times 16$ experiments of SD2.1 and SDXL. From top to bottom, the prompts used in the four groups of examples are: 1. ``A cosmic traveler, floating in zero gravity, spacesuit reflecting the Earth below, stars twinkling in the distance." 2. ``A fierce Viking, axe in hand, leading a raid, the longship slicing through the waves." 3. ``A bustling flower market, stalls filled with bouquets, the air thick with fragrance, people selecting their favorites." 4. ``Tokyo Japan Retro Skyline, Airplane, Railroad Train, Moon etc. - Modern Postcard". Zoom-in for better view.}
\label{fig:sd21_sdxl_visual}

% \vspace{-30pt}
\end{figure*}

\vspace{-5pt}
\subsection{Experimental settings}
\vspace{-5pt}

\textbf{Implementation details.} We evaluate FreCaS on three widely-used pre-trained diffusion models: SD2.1~\citep{ldm}, SDXL~\citep{sdxl} and SD3~\citep{sd3}.
%
% To assess the scalability of our method, 
The sizes of generated images are $\times 4$ and $\times 16$ the original training size. Specifically, we generate images of 1024$\times$1024 and 2048$\times$2048 for SD2.1, while $2048\times 2048$ and $4096\times 4096$ for SDXL.
For SD3, we only generate images of $2048\times 2048$ due to the GPU memory limitation.
We randomly select 10K, 5K, and 1K prompts from the LAION5B aesthetic subset for generating images of 1024$\times$1024, 2048$\times$2048, and 4096$\times$4096, respectively.
We follow the default settings and perform a 50-step sampling process with DDIM sampler for SD2.1 and SDXL, and perform a 28-step sampling process with a flow matching based Euler solver for SD3.
% A 50-step sampling process is adopted with the default sampler (DDIM~\citep{ddim} for SD2.1 and SDXL, and a flow matching based Euler solver ~\citep{rectifiedflow,rectifiedfm} for SD3).
%
For $\times 4$ experiments, we employ two sampling stages at the training size and target size, respectively.
For $\times 16$ experiments, we employ three sampling stages at the training size, $4\times$ training size and $16\times$ training size, respectively.
%
% We set $w_l=7.5$ and $w_h=35$ in FA-CFG. 
% The $w_l$ is decreased to $7$ for SD3 to match its default CFG setting.
%
% For CA-maps reuse, we set $w_c$ as $0.6$.
%
% More details can be found in Appendix~\ref{appendix_setting}.
More details can be found in Appendix \appendixid{experimental_setting}.

\textbf{Evaluation metrics.} \rebuttal{We employ the Fréchet Inception Distance (FID)~\citep{fid} and Inception Score (IS)~\citep{is} to measure the quality of generated images.
Following~\citet{scalecrafter}, we also employ $\text{FID}_b$ as the metric, which is computed on the samples of training size and target size.}
As suggested by~\citet{demofusion}, we report $\text{FID}_p$ and $\text{IS}_p$, which compute the metrics at patch level, to better evaluate the image details.
The CLIP score~\citep{clipscore} is utilized to measure the text prompt alignment of generated images.
As in previous works~\citep{hidiffusion}, we measure the model latency on a single NVIDIA A100 GPU with a batch size of 1. We generate five images and report the averaged latency of the last three images for all methods.
% To evaluate the efficiency of proposed FreCaS, we profile every method on a single NVIDIA A100 GPU and present the latency of generating a image.
\rebuttal{Moreover, we conduct a user study and employ the non-reference image quality assessment metrics to further evaluate our FreCaS. Please refer to Appendix \appendixid{more_metrics} for the details.}

% \textbf{Dataset.} 
% We compare our FreCaS framework with other competing methods on the LAION5B aesthetic subset. 
% Due to concerns regarding potentially illegal content, access to the full LAION-5B dataset has been revoked. 
%
% Specifically, considering the high computational demand, we randomly select 10K, 5K, and 1K prompts from this subset for generating images at resolutions of 1024$\times$1024, 2048$\times$2048, and 4096$\times$4096, respectively.
%
\vspace{-5pt}
\subsection{Experiments on SD2.1 and SDXL}
\vspace{-5pt}
For experiments on SD2.1, we compare FreCaS with DirectInference, MultiDiffusion~\citep{multidiffusion}, AttnEntropy~\citep{attnentropy}, ScaleCrafter~\citep{scalecrafter}, FouriScale~\citep{fouriscale} and HiDiffusion~\citep{hidiffusion}. 
For experiments on SDXL, we compare with DirectInference, AttnEntropy, ScaleCrafter, FouriScale, HiDiffusion, AccDiffusion~\citep{accdiffusion} and DemoFusion~\citep{demofusion}.
\rebuttal{We further compare our FreCaS with training-based methods (Pixart-Sigma~\citep{pixart_sigma} and UltraPixel~\citep{ultrapixel}) and super-resoution methods (ESRGAN~\citep{realesrgan} and SUPIR~\citep{supir}) in Appendix \appendixid{training_SR}.}

\textbf{Quantitative results.} 
Table~\ref{tab:sd21_sdxl} presents quantitative comparisons for $\times 4$ and $\times 16$ generation between FreCaS and its competitors. We can see that FreCaS not only outperforms other methods on synthesized image quality but also exhibits significantly faster inference speed.
\rebuttal{In specific, FreCaS achieves the best FID scores in all experiments of SD2.1 and SDXL, achieving clear advantages over the other methods. In terms of the IS metric, FreCaS performs the best in most cases, only slightly lagging behind DemoFusion on the $\times 16$ experiments of SDXL.}
(Note that DirectInference and MultiDiffusion occasionally achieve higher $\text{FID}_p$ and $\text{IS}_p$ scores because they disregard the issue of duplicated objects.)
For CLIP score, FreCaS obtains the best results on 3 out of the 4 cases, except for the less challenging $\times 4$ generation with SD2.1.

While having superior image quality metrics, FreCaS demonstrates impressive efficiency.
%, especially on generating high-resolution images.
%
It shows more than $2\times$ speedup over DirectInference on $\times 4$ generation experiments, and shows more than $3.6\times$ speedup on the $\times 16$ generation experiments. 
DemoFusion, which is overall the second best method in terms of image quality, is significantly slower than FreCaS. Its latency is about $6\times$ and $7.5\times$ longer than FreCas on $\times 4$ and $\times 16$ experiments, respectively.
On the other hand, HiDiffusion, which is the second faster method, sacrifices image quality for speed.
For example, on the $\times 16$ experiment with SD2.1, HiDiffusion achieves a latency of 18.22s but its $\text{FID}_b$ score is 34.17.
In contrast, FreCaS is faster (13.35s) and has a much better $\text{FID}_b$ score (20.11).
%\footnotetext[1]{\label{footnote_1}\textit{``A cosmic traveler, floating in zero gravity, spacesuit reflecting the Earth below, stars twinkling in the distance.", ``A fierce Viking, axe in hand, leading a raid, the longship slicing through the waves.", ``A bustling flower market, stalls filled with bouquets, the air thick with fragrance, people selecting their favorites." and ``Tokyo Japan Retro Skyline, Airplane, Railroad Train, Moon etc. - Modern Postcard"}}

\textbf{Qualitative results.}
Figure~\ref{fig:sd21_sdxl_visual} illustrates visual comparisons between FreCaS and competitive approaches. From top to bottom are four groups of examples, presenting the results of $\times 4$ generation of SD2.1, $\times 16$ generation of SD2.1, $\times 4$ generation of SDXL, and $\times 16$ generation of SDXL, respectively.
In each group, the top row shows the generated images, while the bottom row shows the zoomed region for better observation.
From Figure~\ref{fig:sd21_sdxl_visual}, we can see that FreCaS effectively synthesizes the described contents while maintaining a coherent scene structure. 
DirectInference, MultiDiffusion and AttnEntropy often produce duplicated objects, such as the many astronauts and warriors. ScaleCrafter and HiDiffusion achieve reasonable image contents in experiments of SD2.1 but generate unnatural layouts in the experiments of SDXL, such as the excessive flowers on the ceiling in $\times 4$ experiment. Our FreCaS consistently maintains coherent image contents and layout in experiments of both SD2.1 and SDXL. AccDiffusion and DemoFusion also achieve natural image contents, but FreCaS generates clearer details such as the flowers and trains.
% Please refer to Appendix~\ref{appendix_visual} for more visual results, including images with other aspect ratios.
Please refer to Appendix \appendixid{more_visual_results} for more visual results, including images with other aspect ratios.

\vspace{-5pt}
\subsection{Experiments on SD3}
\vspace{-5pt}

SD3~\citep{sd3} adopts a rather different network architecture from SD2.1 and SDXL, and many existing methods cannot be applied.
% We can only compare FreCaS with DirectInference and DemoDiffusion~\cite{demofusion}. Due to page limitation, please refer to Appendix~\ref{appendix_sd3} for the results.
% \input{tables/sd3}
We can only compare FreCaS with DirectInference and DemoDiffusion~\cite{demofusion}. Due to page limitation, please refer to Appendix \appendixid{sd3} for the results.

\vspace{-5pt}
\subsection{Ablation Studies}
\vspace{-5pt}

%\rebuttal{In this section, we conduct ablation studies on the $\times 4$ experiments of SDXL to investigate the effectiveness of each components and settings of the FA-CFG strategy, the CA-maps re-utilization and the inference schedule of FreCaS.}

\rebuttal{In this section, we conduct ablation studies on $\times 4$ experiments of SDXL to investigate the effectiveness and settings of our cascaded framework, FA-CFG and CA-maps strategies.}

\begin{figure*}[!t]
\vspace{-15pt}

\centering
\begin{overpic}[width=0.95\linewidth]{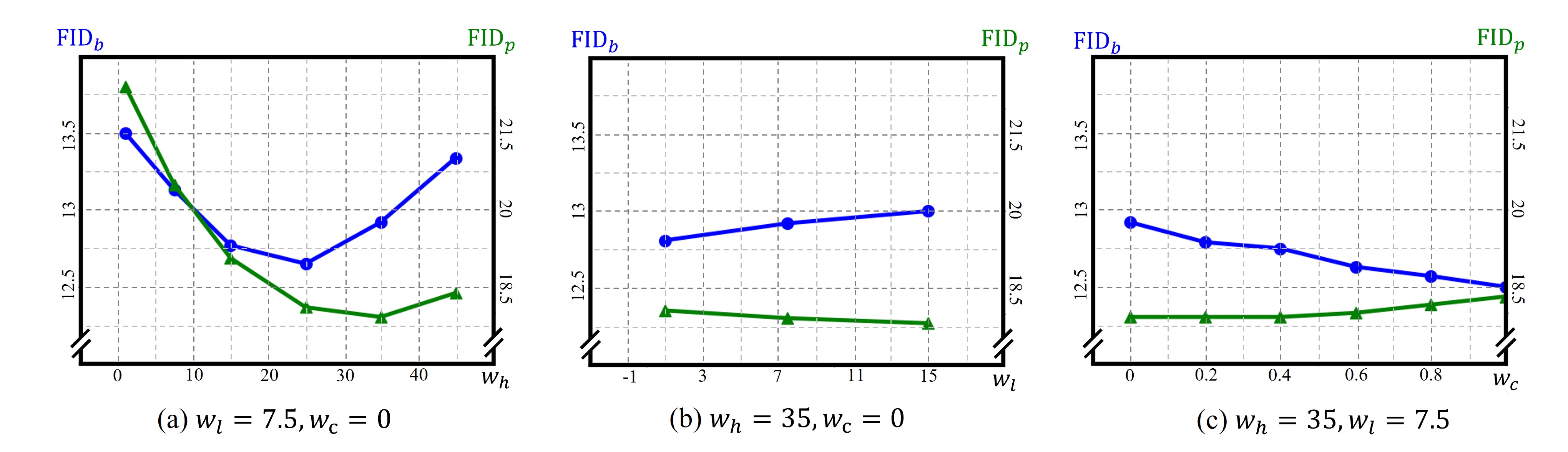}
		%\put(-9,20){\rotatebox{90}{Clip 015/095}}
	\end{overpic}
	
	\vspace{-12pt}
	
	\caption{Ablation studies on $w_l$ and $w_h$ in FA-CFG strategy and $w_c$ in CA-maps reutilization.}
	\label{fig:fig_ab}
\end{figure*}
% There exist two import factors which impact the overall performance of our FreCaS framework.
% The first one is the count of stages for each experiments. 
% For the inference schedule, there are three important factors that control the sampling framework: the timestep of the last state of each stage, the factor for the transition process between stages, and the count of stages. These components are crucial for balancing the trade-off between computational efficiency and the quality of the generated images. Please refer to the Appendix for details.

\begin{figure*}[!t]
\vspace{-5pt}

\begin{overpic}[width=\linewidth]{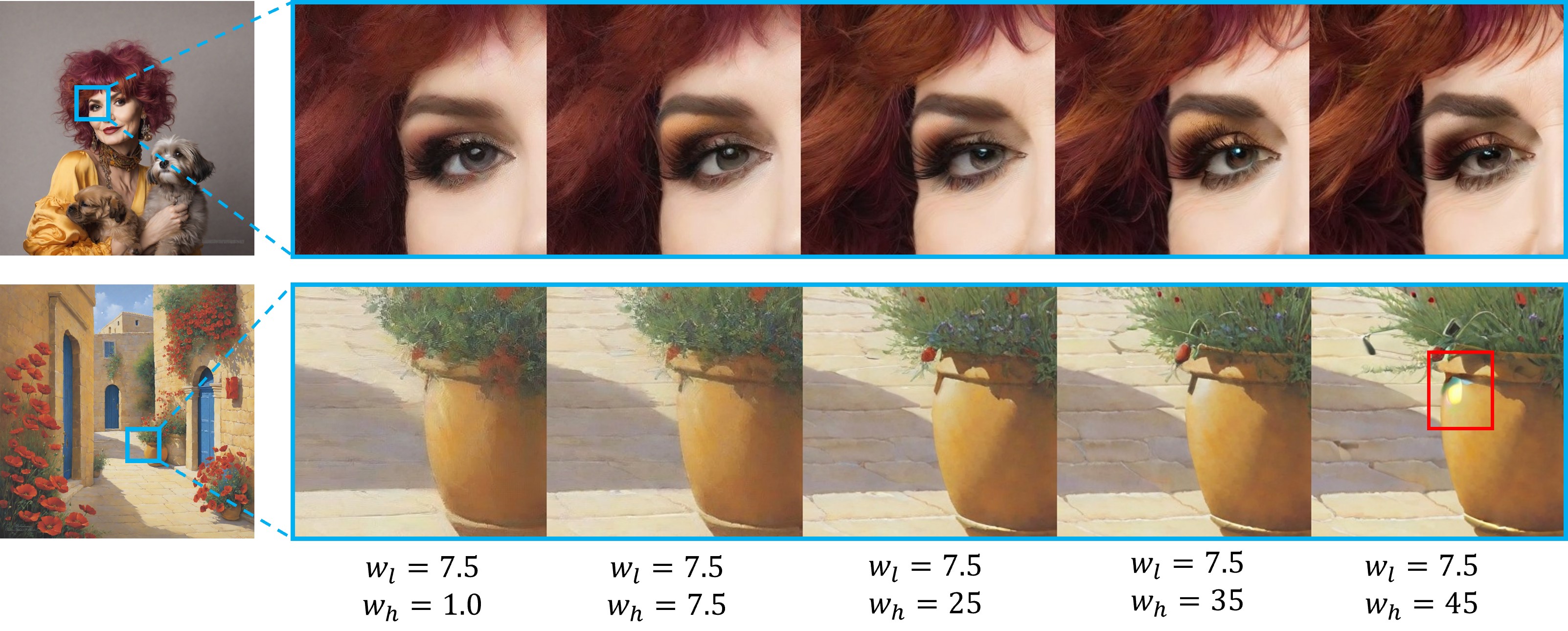}
%\put(-9,20){\rotatebox{90}{Clip 015/095}}
\end{overpic}
\vspace{-20pt}

\caption{Visual results of adjusting $w_h$ in the FA-CFG strategy. From top to bottom, the prompts are ``Eccentric Shaggy Woman with Pet - Little Puppy" and ``Rabat Painting - Mdina Poppies Malta by Richard Harpum", respectively.}
\label{fig:facfg}

\vspace{-15pt}
\end{figure*}
\begin{figure*}[!t]
\vspace{-7pt}
\centering
\begin{overpic}[width=\linewidth]{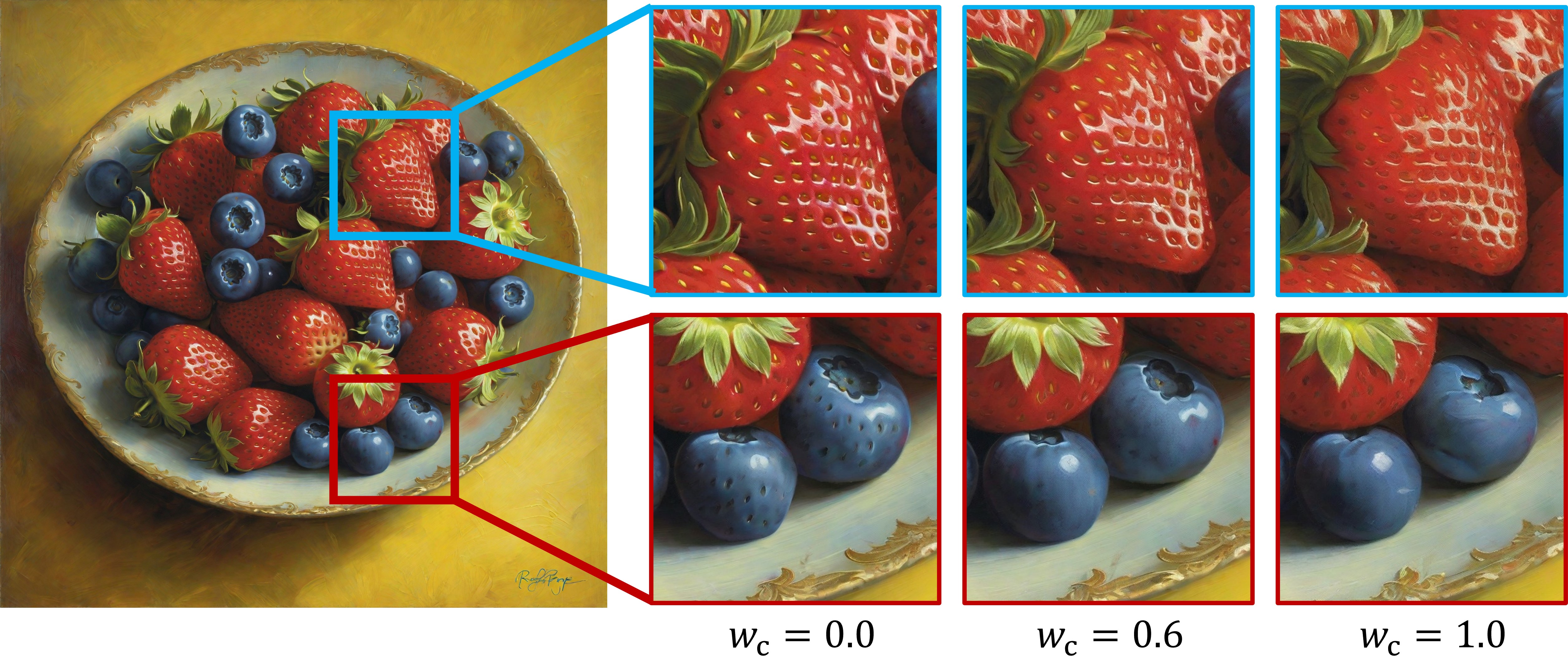}
%\put(-9,20){\rotatebox{90}{Clip 015/095}}
\end{overpic}
\vspace{-23pt}

\caption{Visual results on adjusting $w_c$ in CA-maps reutilization. The prompt is ``Blueberries and Strawberries Art Print".}
\label{fig:cam}

\vspace{-16pt}
\end{figure*}

\textbf{Effectiveness of each component.}
\rebuttal{We conduct a series of ablation studies to better demonstrate the effectiveness of each component of FreCaS, including the cascaded sampling framework, FA-CFG and CA-maps reutilization strategies. Please refer to Appendix \appendixid{each_component} for more details.}

\textbf{FA-CFG strategy.}
The FA-CFG strategy aims to guide the model to generate content within the expanded frequency band. 
To achieve this, FA-CFG introduces two parameters, $w_l$ and $w_h$, to adjust the guidance strength on the low and high frequency components, respectively. When $w_l=w_h$, the FA-CFG strategy degenerates to the conventional CFG approach.
% We conduct a series of experiments to explore the optimal settings for these parameters. The corresponding results are summarized in Table~\ref{tab:facfg} and illustrated Figure~\ref{fig:facfg}.
We conduct a series of experiments to explore the optimal settings of the two parameters. %The results are illustrated in Figure~\ref{fig:fig_ab} and Figure~\ref{fig:facfg}.
%We explore the effect of varying $w_l$ and $w_h$ independently. 
First, we fix $w_l$ at 7.5 and vary $w_h$. The results are shown in Figure~\ref{fig:fig_ab}(a). We observe that as $w_h$ increases from 1.0 to 45, the $\text{FID}_b$ and $\text{FID}_{p}$ metrics initially decrease, indicating improved image quality. However, as $w_h$ becomes too high, the metrics begin to deteriorate. The sweet spot lies between 25 and 35, achieving a low $\text{FID}_b$ of nearly 12.65 and a low $\text{FID}_{p}$ of 17.91.
We then fix $w_h$ at 35 and vary $w_l$. The results are presented in Figure~\ref{fig:fig_ab}(b). Reducing $w_l$ below 7.5 leads to a slight increase in $\text{FID}_p$ from 17.91 to 18.06, whereas increasing $w_l$ over 7.5 deteriorates $\text{FID}_r$ from 12.81 to 13.00. 
Compared to $w_h$, adjusting $w_l$ brings much smaller effects on those two metrics. Thus, we set $w_l$ to 7.5 for experiments on SD2.1 and SDXL, and set it to 7.0 for SD3.
% This suggests that neither extremely high nor low values of $w_l$ are optimal for image quality.

Figure~\ref{fig:facfg} provides visual examples of adjusting $w_h$. Increasing $w_h$ enhances the sharpness of details, such as clearer hair strands and more vivid flower petals. However, an excessively high value of $w_h$ (\eg, 45) will introduce artifacts, as highlighted by the red boxes in the figure. This underscores the importance of selecting an appropriate $w_h$ value to strike a balance between detail enhancement and artifact suppression.
Based on these findings, we set $w_l$ to 7.5 and $w_h$ to 35 yields favorable outcomes in most of the cases.

%\footnotetext[2]{\textit{``Rabat Painting - Mdina Poppies Malta by Richard Harpum" and ``Eccentric Shaggy Woman with Pet - Little Puppy"}}
	
%\footnotetext[3]{\label{footnote_3}\textit{``Blueberries and Strawberries Art Print"}}

\textbf{CA-maps re-utilization.}
To evaluate the effect of weight $w_c$ in the re-utilization of CA-maps, we conduct an ablation study by varying $w_c$ from 0 to 1. The results are shown in Figure~\ref{fig:fig_ab}(c). Increasing $w_c$ continuously decreases $\text{FID}_b$ but increases $\text{FID}_p$, indicating an improvement on the image layout but a drop on image details. To balance between the two metrics, we set $w_c=0.6$. A visual example is shown in Figure~\ref{fig:cam}. We see that this setting leads to a clearer textures on strawberry compared to $w_c=1.0$ and prevents the unreasonable surface of the blueberry in $w_c=0.0$.

\textbf{Inference schedule.}
FreCaS uses two factors to adjust the inference schedule.
The first one is the count of additional stages $N$.
The second factor is the timestep $L$ of last latent in each stage.
% We conduct a series of experiments on the selection of these two factors. The details can be found in Appendix~\ref{appendix_ab}.
We conduct experiments on the selection of these two factors. The details can be found in Appendix \appendixid{inference_schedule}.
Based on results, we set $L$ to $200$, and set $N$ to 2 for $\times 4$ experiments and 3 for $\times 16$ experiments.

\vspace{-5pt}

\section{Conclusion}
\vspace{-5pt}

We developed a highly efficient \textbf{Fre}quency-aware \textbf{Ca}scaded \textbf{S}ampling framework, namely \textbf{FreCaS}, for training-free higher-resolution image generation.
FreCaS leveraged the coarse-to-fine nature of latent diffusion process, reducing unnecessary computations in processing premature high-frequency details.
Specifically, we divided the entire sampling process into several stages having increasing resolutions and expanding frequency bands, progressively generating image contents of higher frequency details.
We presented a \textbf{F}requency-\textbf{A}ware \textbf{C}lassifier-\textbf{F}ree \textbf{G}uidance (\textbf{FA-CFG}) strategy to enable diffusion models effectively adding details of the expanded frequencies, leading to clearer textures.
In addition, we fused the cross-attention maps of previous stages and current one to maintain consistent image layouts across stages.
FreCaS demonstrated advantages over previous methods in both image quality and efficiency. In particular, with SDXL, it can generate a high quality $4096 \times 4096$ resolution image in 86 seconds on an A100 GPU.

\bibliography{iclr2025_conference}
\bibliographystyle{iclr2025_conference}

\ifonepdf
  \pagebreak
  \appendix

\begin{table*}[!t]
	\centering
	\Large
	Appendix to ``FreCaS: Efficient Higher-Resolution Image Generation via Frequency-aware Cascaded Sampling"
	
\end{table*}

In this appendix, we provide the following materials:

\begin{itemize}[leftmargin=1em]
	\item[\ref{appendix_transition}] Details of timestep shifting in the transition process (referring to Sec. 3.2 in the main paper); 
	\item[\ref{appendix_setting}] The detailed settings of FreCaS on $\times 4$ and $\times 16$ generation for SD2.1, SDXL and SD3 (referring to Sec. 4.1 in the main paper);
\item[\ref{more_metrics}] \rebuttal{Results of user studies and non-reference image quality assessment (NR-IQA) (referring to Sec. 4.1 in the main paper);}
\item[\ref{training_SR}] \rebuttal{Comparison with training-based methods and super-resolution methods (referring to Sec. 4.2 in the main paper);}
\item[\ref{appendix_visual}] \rebuttal{More visual results and visual comparisons (referring to Sec. 4.2 in the main paper);}
\item[\ref{appendix_sd3}] Experimental results of $\time 4$ generation of SD3 (referring to Sec. 4.3 in the main paper);
\item[\ref{appendix_ab}] \rebuttal{Ablation studies on individual components of FreCaS and inference schedule (referring to Sec. 4.4 in the main paper).}
\end{itemize}

\section{Shifting Timestep in the Transition Process}
\label{appendix_transition}

% In Sec.\ref{transition_process}, 
As mentioned in Sec. 3.2 of the main paper, FreCaS employs a five-step transition process to transform the last latent in the current stage \(\vz^{s_{i-1}}_{L}\) to the first latent in the next stage \(\vz^{s_i}_{F}\).
In addition to changing the resolution, we adjust the timestep from $L$ to $F$ to ensure that the signal-to-noise ratio (SNR) \citep{SNR} could be a constant in the transition process.
Given a state $\vz_t$ = $\sqrt {\alpha_t} \vz_\text{0} + \sqrt {1-\alpha_t} \epsilon$ at timestep $t$, the SNR is defined as $\text{SNR}(\vz_t) = \frac{\alpha_t}{1-\alpha_t}$,
where \(\alpha_1, \ldots, \alpha_T\) represent the noise schedule, and \(\epsilon\) is Gaussian noise.
It has been found \citep{simplediffusion, importance} that the SNR maintains a consistent ratio across resolutions for diffusion models using the same noise schedule:
\[ \text{SNR}(\vz_t^{s}) = \text{SNR}(\vz_t^{\hat s}) \cdot \left( \frac{s}{\hat s} \right)^\gamma, \]
where \( s \) and \( \hat s \) denote different resolutions. The value of \(\gamma\) is typically set to 2.

\cite{relay} and \cite{fdm} proposed to redesign the noise schedule to keep SNR consistent when changing the resolutions of intermediate states. Since the pre-trained diffusion models have fixed noise schedules, in this paper we adjust the timestep, instead of the noise schedule, to ensure consistent SNR between \(\vz_{L}^{s_{i-1}}\) and \(\vz_{F}^{s_{i}}\):
\begin{equation} 
	\text{SNR}(\vz_{{L}}^{s_{i-1}}) = \text{SNR}(\vz_{{F}}^{s_{i}}) \Rightarrow F = \alpha^{-1} \left( \frac{\left( \frac{s_{i-1}}{s_{i}} \right)^\gamma \cdot \alpha_{{L}}}{1 + \left( \left( \frac{s_{i-1}}{s_i} \right)^\gamma - 1 \right)\cdot \alpha_{{L}}} \right),
	\end{equation}
where \(\alpha^{-1}\) is the inverse function of \(\alpha_t\). Proper adjustment of \(\gamma\) can yield additional improvements.

% 关于SD3的公式，个人感觉还是必要的。其和SD21/SDXL的公式有细微不同，也不没有$\gamma$参数可调，在table1中有明显区别
Besides, SD3~\citep{sd3} employs a similar formula to shift the timestep when varying resolutions:
\begin{equation}
	F = \frac {\sqrt {\frac {s_{i}}{s_{i-1}}} \cdot L}{1+(\sqrt {\frac {s_{i}}{s_{i-1}}}-1)\cdot {L}}.
\end{equation}

\section{Experimental Setting Details}
\label{appendix_setting}
The experimental setting details of our FreCaS are listed in Table~\ref{tab:appendix_settings}.

\begin{table}[t]
	\centering
	\renewcommand{\arraystretch}{1.5}
    \vspace{-10pt}
	
	\caption{Detailed settings of FreCaS on the experiments. $N$ denotes the count of additional stages. ``Steps" presents the sampling steps in each stage. $L$ presents the timestep of last latent in each stage except for the final one. $\gamma$ denotes the SNR ratio in the transition process. $w_l$, $w_h$ and $w_c$ are the hyper-parameters of the proposed FA-CFG and CA-maps re-utilization.}
	
	\label{tab:appendix_settings}
	\vspace{5pt}
	
	\begin{tabular}{|c|c|cccc|ccc|}
	\hline
		
    \diagbox{}{} & \diagbox{}{} & $N+1$ & Steps & $L$ & $\gamma$ & $w_l$ & $w_h$ & $w_c$\\
\hline
\hline

\multirow{2}*{SD2.1} & $\times$4 & 2 & 40,10 & 100 & 3.0 & 7.5 & 45.0 & 0.6 \\
\cline{2-9}
& $\times$16 & 3 & 30,10,10 & 200,200 & 3.0 & 7.5 & 35.0 & 0.4 \\
\hline

\multirow{2}*{SDXL} & $\times$4 & 2 & 40,10 & 200 & 1.5 & 7.5 & 35.0 & 0.6 \\
\cline{2-9}
&  $\times$16 & 3 & 30,5,15 & 400,200 & 2.0 & 7.5 & 35.0 & 0.6 \\
\hline

\multirow{1}*{SD3} & $\times$4 & 2 & 20,8 & 50 & - & 7.0 & 35.0 & 0.5 \\
\hline

\hline

	\end{tabular}

\end{table}

\section{Results of User studies and NR-IQA Metrics}
\label{more_metrics}
\rebuttal{We have (a) conducted user studies and (b) employed non-reference image quality assessment (NR-IQA) metrics to further assess the performance of FreCaS and its competing methods.}

\begin{figure}
\centering
\includegraphics[width=0.4\linewidth]{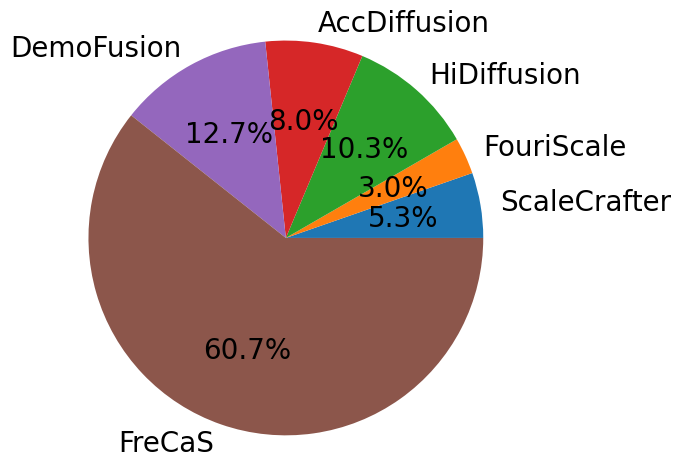}
    \caption{\rebuttal{User study results on $\times 4$ generation of SDXL.}}
    \label{fig:user_study}
\end{figure}

\subsection{User Studies}
\rebuttal{For the user studies, we compare FreCaS with ScaleCrafter, FouriScale, HiDiffusion, DemoFusion, and AccDiffusion on 2048$\times$2048 image generation using SDXL. We randomly selected 30 prompts and generated one image per method for each prompt, creating 30 sets of images. Ten volunteers participated in the test, and they were asked to select the image with the best details and reasonable semantic layout from each set. The results are shown in Figure~\ref{fig:user_study}. We can see that FreCaS significantly outperforms other methods, with 60\% votes as the best method. DemoFusion, AccDiffusion, and HiDiffusion perform similarly, with each having about 10\% of the votes. In contrast, FouriScale and ScaleCrafter have the fewest votes, about 5\% each.}

\begin{table*}[!t]
\centering
\renewcommand{\arraystretch}{1.2}
\caption{\rebuttal{NR-IQA metrics on $\times 4$ and $\times 16$ generation of SDXL.\\}}

\label{tab:iqa}

\begin{tabular}{|c|c|c|c|c|c|c|}    
    \hline
     \multirow{2}*{Methods} & \multicolumn{3}{c|}{$\times$4} & \multicolumn{3}{c|}{$\times$16} \\
     \cline{2-7}
     & \footnotesize{\text{clipiqa}}$\uparrow$ & \footnotesize{\text{niqe}}$\downarrow$ & \footnotesize{musiq}$\uparrow$ & \footnotesize{\text{clipiqa}}$\uparrow$ & \footnotesize{\text{niqe}}$\downarrow$ & \footnotesize{musiq}$\uparrow$ \\
    \hline

DirectInference & 0.522 & 4.167 & 53.98 & 0.469 & 4.370 & 29.00\\
AttnEntropy & 0.547 & 4.210 & 54.87 & 0.528 & 4.614 & 27.98 \\
ScaleCrafter & 0.664 & 3.577 & 61.12 & 0.618 & 3.783 & 36.00 \\
FouriScale & 0.662 & 3.580 & 60.77 & 0.612 & 3.791 & 35.52\\
HiDiffusion & \best{0.690} & 4.049 & 61.69 & 0.574 & 7.348 & 36.71 \\
AccDiffusion & 0.627 & 3.641 & 57.02 & 0.626 & 3.587 & 31.83 \\
DemoFusion & 0.651 & 3.410 & 58.98 & 0.637 & 3.376 & 33.46\\
\colorbox{gray!30}{\textbf{\qquad Ours\qquad}} & 0.668 & \best{3.391} & \best{63.10} & \best{0.646} & \best{3.367} & \best{37.33}\\
    \hline

\end{tabular}
\end{table*}
\subsection{NR-IQA Metrics}
\rebuttal{For the NR-IQA metrics, we employ CLIPIQA~\citep{clipiqa}, NIQE~\citep{niqe}, and MUSIQ~\citep{musiq} on $\times4$ and $\times 16$ image generations with SDXL. The results are presented in Table~\ref{tab:iqa}. Our FreCaS consistently outperforms all the other methods. For example, on $\times 4$ generation, FreCaS achieves a CLIPIQA score of 0.668, a NIQE score of 3.391, and a MUSIQ score of 63.10, compared to 0.651, 3.410, and 58.98 for DemoFusion. On $\times 16$ generation, FreCaS achieved a CLIPIQA score of 0.646, a NIQE score of 3.367, and a MUSIQ score of 37.33, compared to 0.626, 3.587, and 31.83 for AccuDiffusion. Notably, FreCaS only lags behind HiDiffusion on the CLIPIQA metric in $\times 4$ image generation.}

\section{Comparison with training-based methods and super-resolution methods}
\label{training_SR}

\modifyCR{\subsection{quantitative and visual comparison}}

\begin{figure*}
	\centering
	\includegraphics[width=\linewidth]{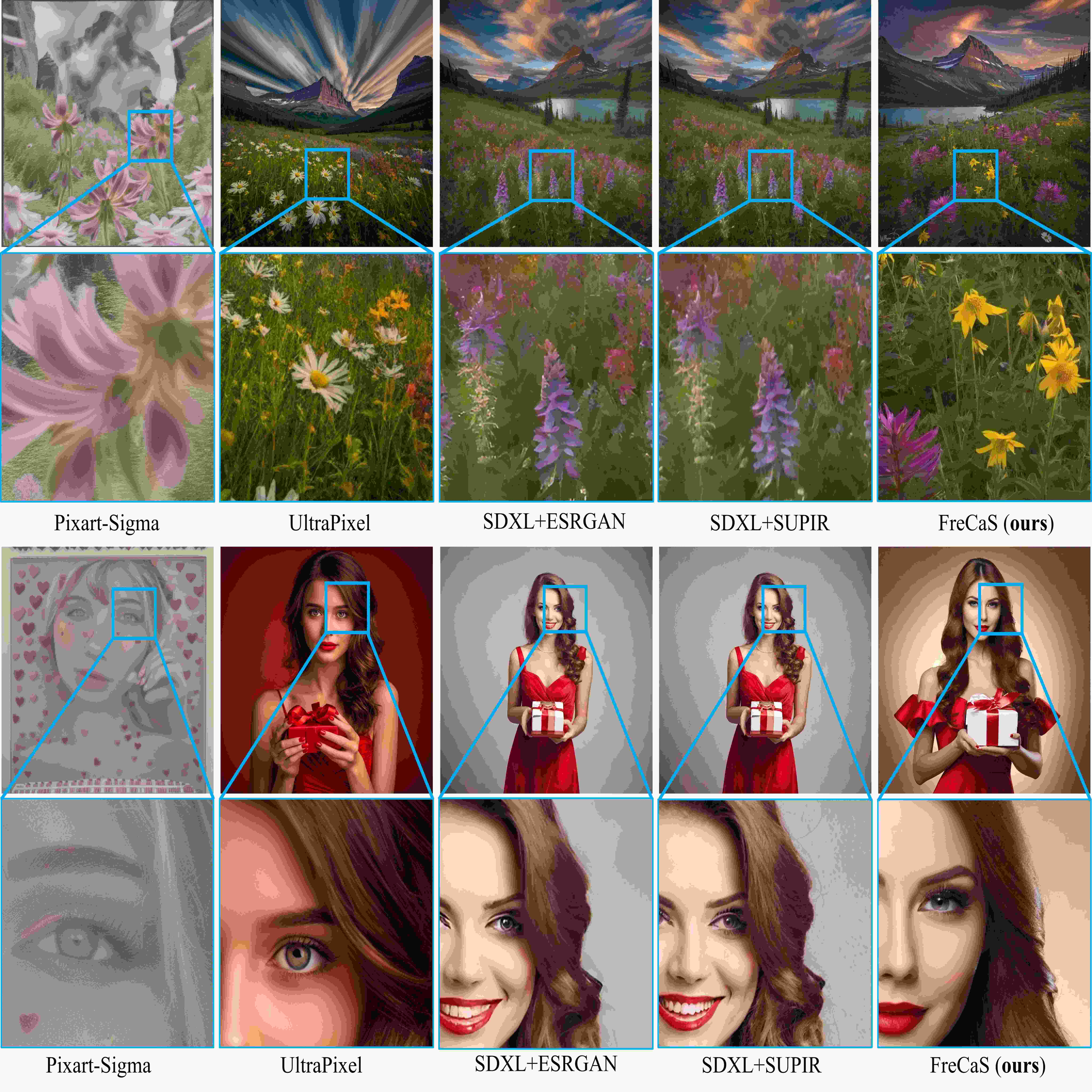}
	\caption{\rebuttal{Visual comparison with training-based methods and super-resolution methods on $\times 4$ generation of SDXL.}}
	\label{fig:training_sr}
\end{figure*}

\rebuttal{We conducted additional experiments comparing FreCaS with training-based methods (Pixart-Sigma~\citep{pixart_sigma} and UltraPixel~\citep{ultrapixel}) and super-resolution methods (ESRGAN~\citep{realesrgan} and SUPIR~\citep{supir}). 
To ensure fair comparisons, we set the model precision to fp16 (bf16 for UltraPixel, as recommended by the authors) and use the DDIM sampler for diffusion-based methods. For Pixart-Sigma, we can only report its results for 2048$\times$2048 image generation since its 4K model is not available. The quantitative results are summarized in Table~\ref{tab:other_method}.}

\rebuttal{From Table~\ref{tab:other_method}, we can see that FreCaS outperforms Pixart-Sigma and UltraPixel in most metrics. For example, FreCaS achieves an FID score of 16.48 and an IS score of 17.18, compared to 26.1 and 14.44 of Pixart-Sigma, and 25.56 and 17.11 of UltraPixel on the $\times 4$ image generation task. This is because Pixart-Sigma, as acknowledged by the authors, heavily relies on the advanced samplers (see \href{https://github.com/PixArt-alpha/PixArt-sigma/issues/65}{https://github.com/PixArt-alpha/PixArt-sigma/issues/65}) so that the results are not very stable. UltraPixel, while achieving comparable performance to DemoFusion, still lags behind FreCaS in most metrics. Besides, the two methods are much slower than our FreCaS.}

\rebuttal{For SR-based methods, FreCaS may have lower FID, IS, and CLIP scores than SDXL+ESRGAN. This is because SR methods are designed to strictly adhere to low-resolution inputs, while these metrics (FID, IS, and CLIP) evaluate images by downsampling them to low resolution, which cannot well reflect the quality of generated high-resolution images. However, FreCaS significantly outperforms SDXL+ESRGAN in $\text{FID}_p$ and $\text{IS}_p$. Specifically, FreCaS achieves an $\text{FID}_p$ score of 39.82 and an $\text{IS}_p$ score of 14.16, compared to 43.10 and 13.48 of SDXL+ESRGAN on $\times 16$ image generation.
This indicates its superior ability to generate high-resolution local details. This observation is consistent with the findings in the DemoFusion paper. Additionally, SDXL+SUPIR outperforms FreCaS in $\text{FID}_p$ and $\text{IS}_p$, but at the cost of much longer inference latency (85.87 seconds for FreCaS vs. 512.4 seconds for SDXL+SUPIR on $\times 16$ image generations).}

\rebuttal{We have provided some visual comparisons in Figure~\ref{fig:training_sr}. One can see that FreCaS demonstrates better visual quality than either training-based or SR-based methods in high-resolution image generation, such as the more vivid and clearer flowers, hairs and the more natural color of lips.}

\begin{table*}[!t]
\centering
\renewcommand{\arraystretch}{1.2}
\caption{\rebuttal{Comparison with training-based methods and super-resolution methods on $\times 4$ and $\times 16$ generation of SDXL.\\}}

\label{tab:other_method}

\begin{tabular}{|c|c|c|c|c|c|c|c|c|c|c|c|c|c|c}    
    \hline
     & Methods & \footnotesize{$\text{FID}$}$\downarrow$ & \footnotesize{$\text{FID}_p$}$\downarrow$ & \footnotesize{IS}$\uparrow$ & $\text{\footnotesize{IS}}_p$$\uparrow$ & \makecell{CLIP\\SCORE}$\uparrow$ & \footnotesize{Latency}(s)$\downarrow$ \\
    \hline

 \multirow{5}*{\rotatebox{90}{$\times$4}}  
 %& Pixart-Sigma (DPM) & 19.98 & 27.69 & 15.60 & 14.11 & 30.38 & 71.56 \\
 & Pixart-Sigma & 26.11 & 38.58 & 14.44 & 14.45 & 28.10 & 71.45 \\
    & UltraPixel & 25.56 & 19.95 & 17.11 & 17.10 & 33.17 & 41.70\\
    & SDXL+ESRGAN & 13.03 & 18.10 & 17.30 & 16.58 & 34.13 & 6.36 \\
    & SDXL+SUPIR & 12.08 & 17.31 & 17.57 & 17.12 & 34.16& 105.5 \\
    & \colorbox{gray!30}{\textbf{\qquad Ours\qquad}} & 16.48 & 17.91 & 17.18 & 17.31 & 33.28 & 13.84 \\
    \cline{1-12}
\multirow{4}*{\rotatebox{90}{$\times$16}} 
    & UltraPixel & 51.43 & 45.88 & 12.48 & 13.73 & 33.07 & 162.4 \\
    & SDXL+ESRGAN & 45.86 & 43.10 & 12.94 & 13.48 & 33.44 & 7.25 \\
    & SDXL+SUPIR & 43.94 & 39.35 & 13.22 & 14.37 & 33.49 & 512.4 \\
    & \colorbox{gray!30}{\textbf{\qquad Ours\qquad}} & 42.75 & 39.82 & 12.68 & 14.16 & 33.03 & 85.87 \\
    \hline

\end{tabular}

\end{table*}

\modifyCR{\subsection{Stability Metrics}}

\begin{table*}
	\centering
	\renewcommand{\arraystretch}{1.2}
	\caption{\modifyCR{Stability experiments on 200 images of 20 prompts on $\times4$ generation. 
			%``AoS" denotes average of standard deviations of each prompt. ``SoM" denotes the standard deviations of mean metrics across all 20 prompts. Please refer to the main response for more details.
	}}
	\label{tab:stability}
	\begin{tabular}{|c|c|c|c|c|c|c|c|c|c|}    
		\hline
		\multirow{2}*{Methods} & \multicolumn{3}{c|}{clipiqa$\uparrow$} & \multicolumn{3}{c|}{niqe$\downarrow$} & \multicolumn{3}{c|}{musiq$\uparrow$} \\
		\cline{2-10}
		& Mean & AoS  & SoM & Mean & AoS & SoM & Mean & AoS & SoM \\
		\hline
		Pixart-Sigma & 0.558 & 0.05 & 0.11 & 5.256 & 0.35 & 0.95 & 51.546 & 4.49 & 8.24 \\
		UltraPixel  & 0.540 & 0.04 & 0.11 & 4.625 & 0.42 & 1.55 & 56.215 & 2.94 & 7.95 \\
		FreCaS & 0.633 & 0.11 & 0.04 & 3.886 & 0.87 & 0.27 & 59.756 & 9.64 & 2.95 \\
		\hline
		
		\hline
	\end{tabular}
\end{table*}

\modifyCR{To quantitatively analyze the stability of training-based and training-free methods, we generated 200 images for 20 randomly selected prompts (10 images for each prompt) using Pixart-Sigma (with default sampler unless otherwise stated), UltraPixel, and our FreCaS. Considering that FID and IS are not suitable for evaluating individual examples, we adopt the NR-IQA metrics (CLIPIQA, NIQE, and MUSIQ) to measure the performance of each method. In specific, we define the following three measures to evaluate the generation quality, stability and consistency of each method.}
\begin{itemize}
	\item \textbf{Average Score (Mean):} \modifyCR{The average score across the 200 generated images for each of the three metrics (CLIPIQA, NIQE, and MUSIQ). This metric can reflect the generation quality of each method.}
	\item \textbf{Average of Standard Deviations (AoS):} \modifyCR{We first compute the standard deviation of the metrics for each prompt across 10 runs, and then report the average of these standard deviations across all 20 prompts. This metric can reflect the stability of each method.}
	\item \textbf{Standard Deviation of Averages (SoM):} \modifyCR{We first compute the mean of the metrics for each prompt across 10 runs, and then report the standard deviation of these mean values across all 20 prompts. This metric can reflect the consistency of a method's performance across different prompts.}
\end{itemize}
\modifyCR{The results are listed in Table~\ref{tab:stability}. }
\modifyCR{From this table, we can see that our FreCaS achieves the highest ``Mean" scores across the three metrics, demonstrating the best performance in term of generation quality. Pixart-Sigma and UltraPixel have smaller AoS scores than FreCaS, indicating better stability for the same input prompt.
However, FreCaS demonstrates significantly better SoM scores than Pixart-Sigma and UltraPixel, indicating that it can consistently achieve better results across various prompts.
}

\modifyCR{\subsection{User studies on visual quality and success rate}}

\begin{table*}[!t]
	\centering
	\renewcommand{\arraystretch}{1.2}
	\caption{\modifyCR{\textbf{User studies of visual quality and success rate on $\times 4$ generation with SDXL.\\}}}
	
	\label{tab:us2}
	
	\begin{tabular}{|c|c|c|c|c|}    
		\hline
		\multirow{2}*{Methods} & \multicolumn{2}{c|}{Image Quality} & \multicolumn{2}{c|}{Success Rate} \\
		\cline{2-5}
		& Counts & Percentage & Counts & Percentage \\
		\hline
		Pixart-Sigma & 5 & 5\% & 52 & 20.8\%\\
		UltraPixel & 37 & 37\% & 96 &38.4\% \\
		\colorbox{gray!30}{\textbf{\qquad Ours\qquad}} & 58 & 58\% &68 & 27.2\% \\
		\hline
		
	\end{tabular}
\end{table*}

\modifyCR{ We conducted user studies to explore the generated image quality and success rate of Pixart-Sigma, UltraPixel, and our FreCaS. The results are listed in Table \ref{tab:us2}.}

\modifyCR{For the study on generation quality, we randomly select 20 prompts from Laion5B and generate one image per method for each prompt, creating 20 sets of images (3 images per set). Five volunteers (3 males and 2 females) were invited to participate in the test. All the volunteers are not working in the area of image generation to avoid potential bias. Each time, the set of 3 images for the same prompt are presented to the volunteers in random order. The volunteers can view the images multiple times, and they are asked to select the image with the best quality from each set. There are 100 votes in total.}

\modifyCR{For the study on success rate, we randomly select 10 prompts and generate five images per prompt for each method, resulting in 50 images per method. We invited the same five volunteers as in the study of generation quality to judge whether the generated image is a success or failure. When making the decision, the volunteers are instructed to consider two factors. First, whether the image content is faithful to the description of the prompt. Second, whether the image quality is satisfactory. Only when both the two requirements are met, the generation is considered as a success. There are 250 judges for each method.}

\modifyCR{As we can see from Table \ref{tab:us2}, our FreCaS outperforms significantly Pixart-Sigma and UltraPixel in terms of image generation quality, with 58\% of the images being voted as the best. In terms of success rate, UltraPixel works the best, with 96 out of 250 images being marked as successful. Our FreCaS lags behind, with 68 successful cases. However, our FreCaS still generates more successful results than Pixart-Sigma (52 images), indicating that a well designed training-free method can surpass some training-based methods.}
\modifyCR{Furthermore, we can also observe that none of the methods, including training-based and training-free ones, achieves a success rate higher than 40\%. This implies that there are much space to improve. }

\section{More Visual Results}
\label{appendix_visual}

\begin{figure*}[!t]

% \vspace{-5pt}
	
\begin{overpic}[width=\linewidth]{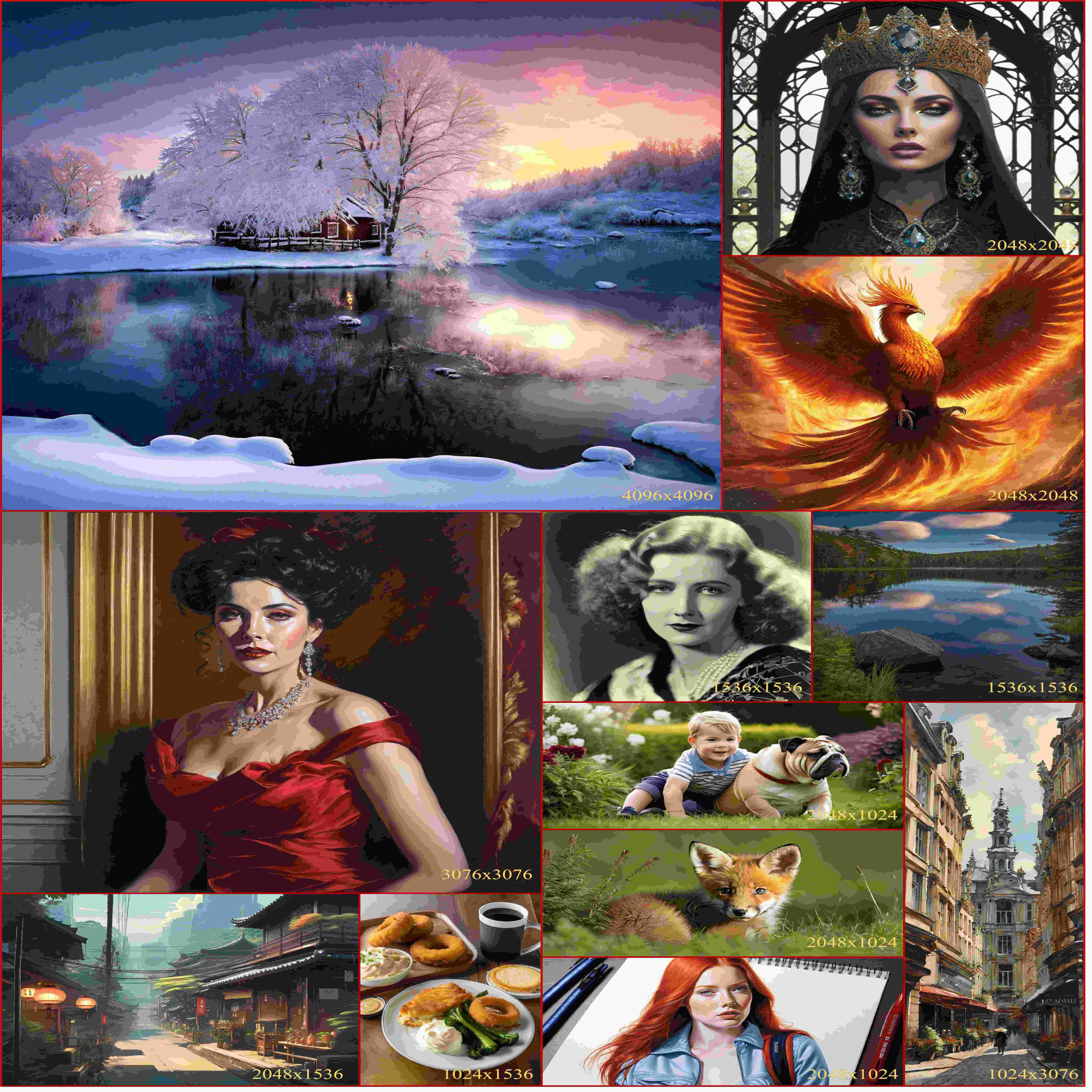}
\end{overpic} 

%\vspace{-5pt}

\caption{Visual results of FreCaS on SDXL. Please zoom-in for better view.}
% From top to bottom, from left to right, the prompts used in examples are: \scriptsize{1. ``Beautiful winter wallpapers." 2. ``A regal queen adorned with jewels." 3. ``A majestic phoenix, wings ablaze, rising from ashes, the flames casting a warm glow." 4. ``Lady in Red oil portrait painting won the John Singer Sargent People's award." 5. ``Star of the day -- Actress Evelyn Laye - 1917." 6. ``Photograph - Clouds Over Daicey Pond by Rick Berk." 7. ``little-boy-with-large-bulldog-in-a-garden-france." 8. ``03-Brussels-Maja-Wronska-Travels-Architecture-Paintings.", 9. ``Red Fox Pup Print by William H. Mullins." 10. ``Lovely Illustrations Of Cityscapes Inspired By Southeast Asia Malaysian digital illustrator Chong Fei Giap's illustrations of cityscapes are lovely and inspiring. Fantasy Landscape, Landscape Art, Illustrator, Japon Tokyo, Animation Background, Art Background, Matte Painting, Anime Scenery, Jolie Photo." 11. ``A plate with creamy chicken and vegetables, a side of onion rings, a cup of coffee and a slice of cheesecake." 12. ``Hyper-Realistic Portrait of Redhead Girl Drawn with Bic Pens."}. Zoom-in for better view.}
\label{fig:appendix_visual_results}

% \vspace{-50pt}

\end{figure*}

\subsection{More Visual Results}
Figure~\ref{fig:appendix_visual_results} illustrates more visual results of FreCaS, including those with varying aspect ratios.
From top to bottom, and left to right, the prompts used in examples are: 1. ``Beautiful winter wallpapers." 2. ``A regal queen adorned with jewels." 3. ``A majestic phoenix, wings ablaze, rising from ashes, the flames casting a warm glow." 4. ``Lady in Red oil portrait painting won the John Singer Sargent People's award." 5. ``Star of the day -- Actress Evelyn Laye - 1917." 6. ``Photograph - Clouds Over Daicey Pond by Rick Berk." 7. ``little-boy-with-large-bulldog-in-a-garden-france." 8. ``03-Brussels-Maja-Wronska-Travels-Architecture-Paintings.", 9. ``Red Fox Pup Print by William H. Mullins." 10. ``Lovely Illustrations Of Cityscapes Inspired By Southeast Asia Malaysian digital illustrator Chong Fei Giap's illustrations of cityscapes are lovely and inspiring. Fantasy Landscape, Landscape Art, Illustrator, Japon Tokyo, Animation Background, Art Background, Matte Painting, Anime Scenery, Jolie Photo." 11. ``A plate with creamy chicken and vegetables, a side of onion rings, a cup of coffee and a slice of cheesecake." 12. ``Hyper-Realistic Portrait of Redhead Girl Drawn with Bic Pens."

\rebuttal{To further validate the performance of FreCaS in real-world application scenarios, we have provided additional visual results in three categories:
\begin{itemize}
	\item \textbf{Simple scenes.} These images typically contain a single object in a realistic style. We display images of people, animals, landscapes, buildings, and other common objects. The visual results for this group are presented in Figure~\ref{fig:simple}. 
	\item \textbf{Various styles.} This group showcases images in different artistic styles, including oil painting, pencil sketch, ink wash, watercolor, and poster art. The results are shown in the first two rows of Figure~\ref{fig:complex}.
	\item \textbf{Complex scenes.} These images contain multiple objects or have intricate textures. The results are displayed in the bottom two rows of are presented in Figure~\ref{fig:complex}.
\end{itemize}
From these visual results, it is evident that FreCaS consistently generates high-quality images across various styles and contents, demonstrating the capability of FreCaS in real-world applications.}

\begin{figure}
	\centering
	\includegraphics[width=1.0\linewidth]{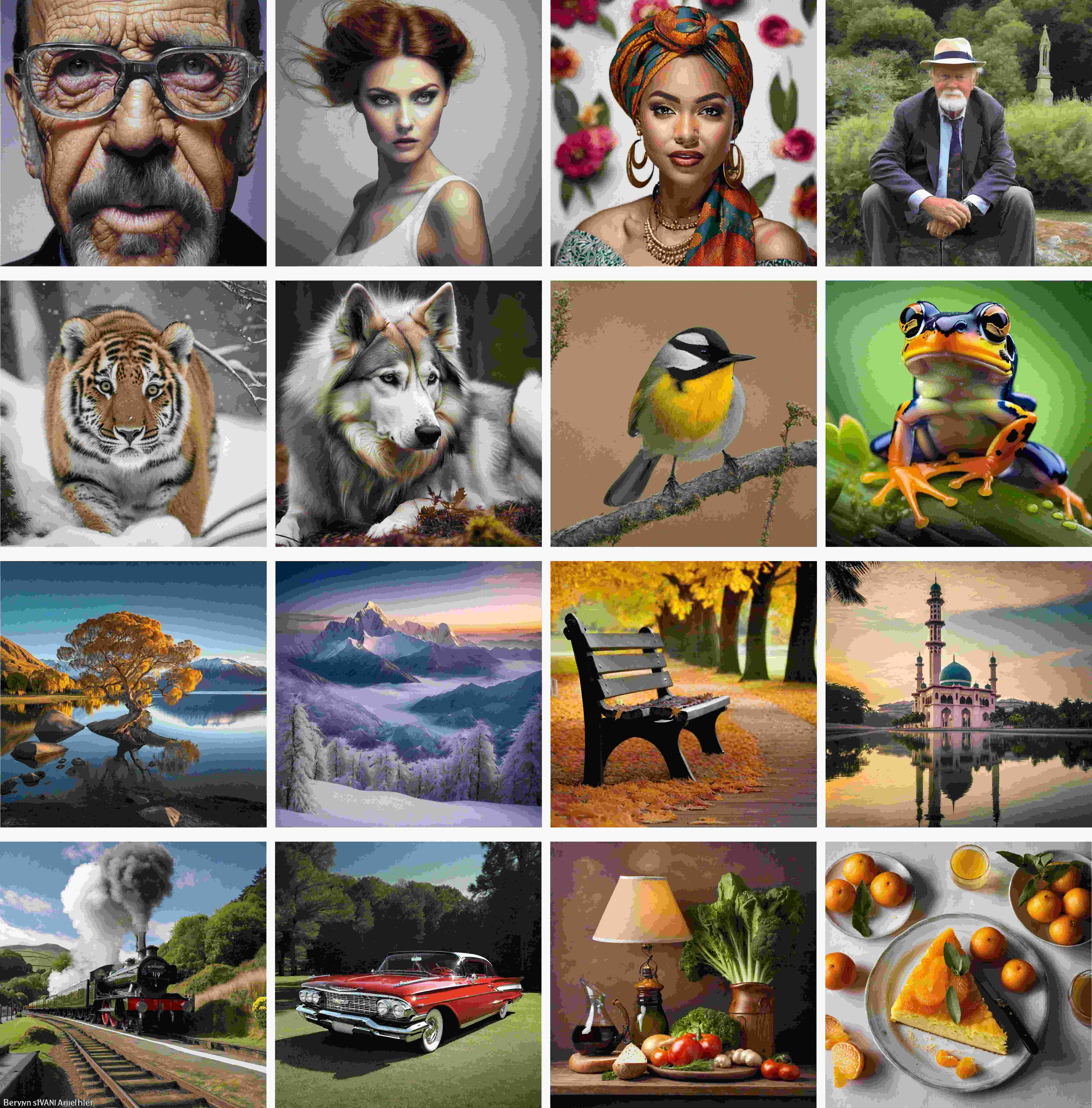}
	\caption{\rebuttal{More visual results on simple scenes.}}
	\label{fig:simple}
\end{figure}

\begin{figure}
	\centering
	\includegraphics[width=1.0\linewidth]{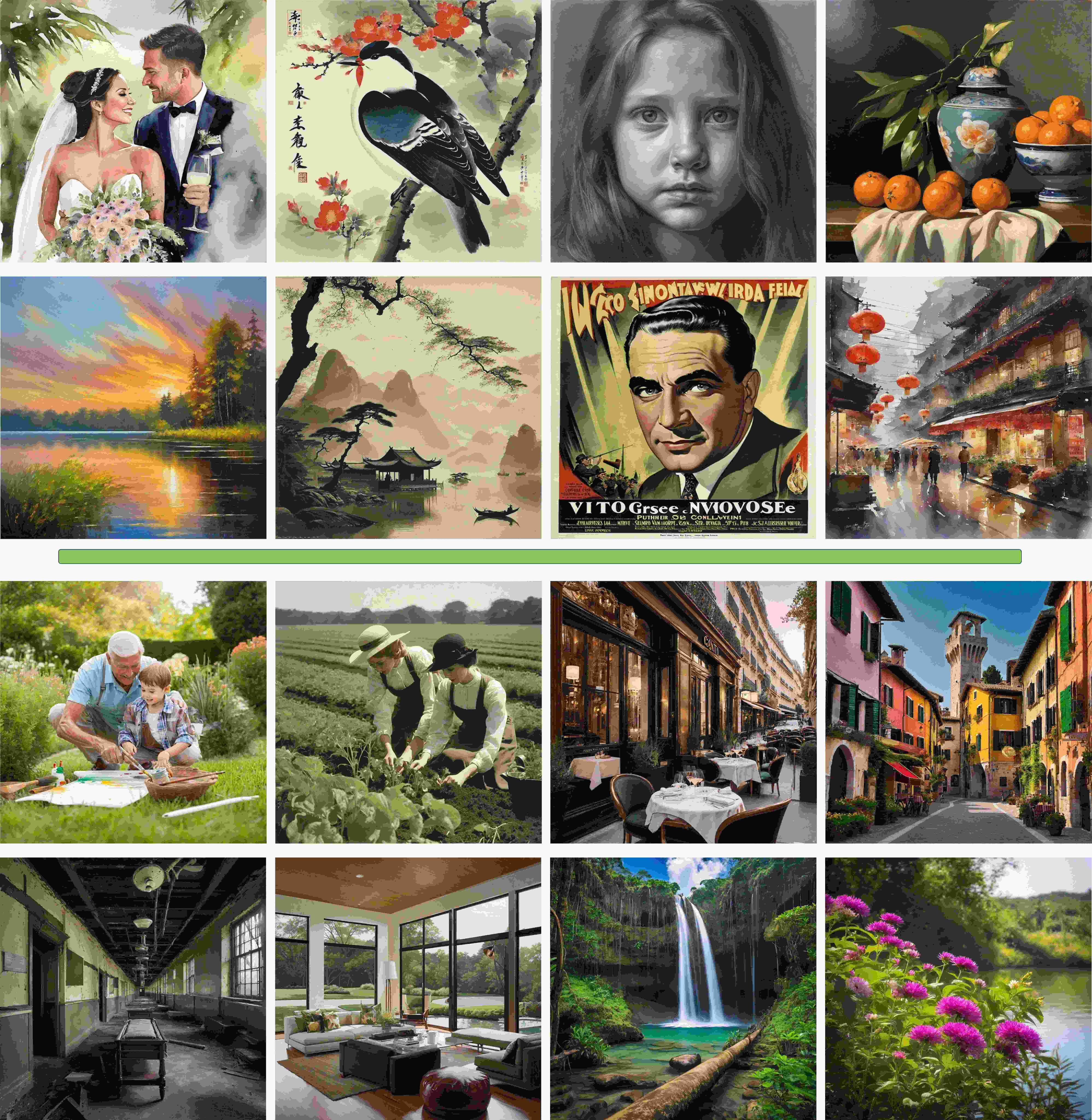}
	\caption{\rebuttal{More visual results of various styles (top two rows) and complex scenes (bottom two rows).}}
	\label{fig:complex}
\end{figure}

\begin{figure*}[!t]

% \vspace{-10pt}
% \begin{overpic}[width=0.97\linewidth]{figures/appendix_visual_comparsion_sd21.jpg}
\begin{overpic}[width=\linewidth]{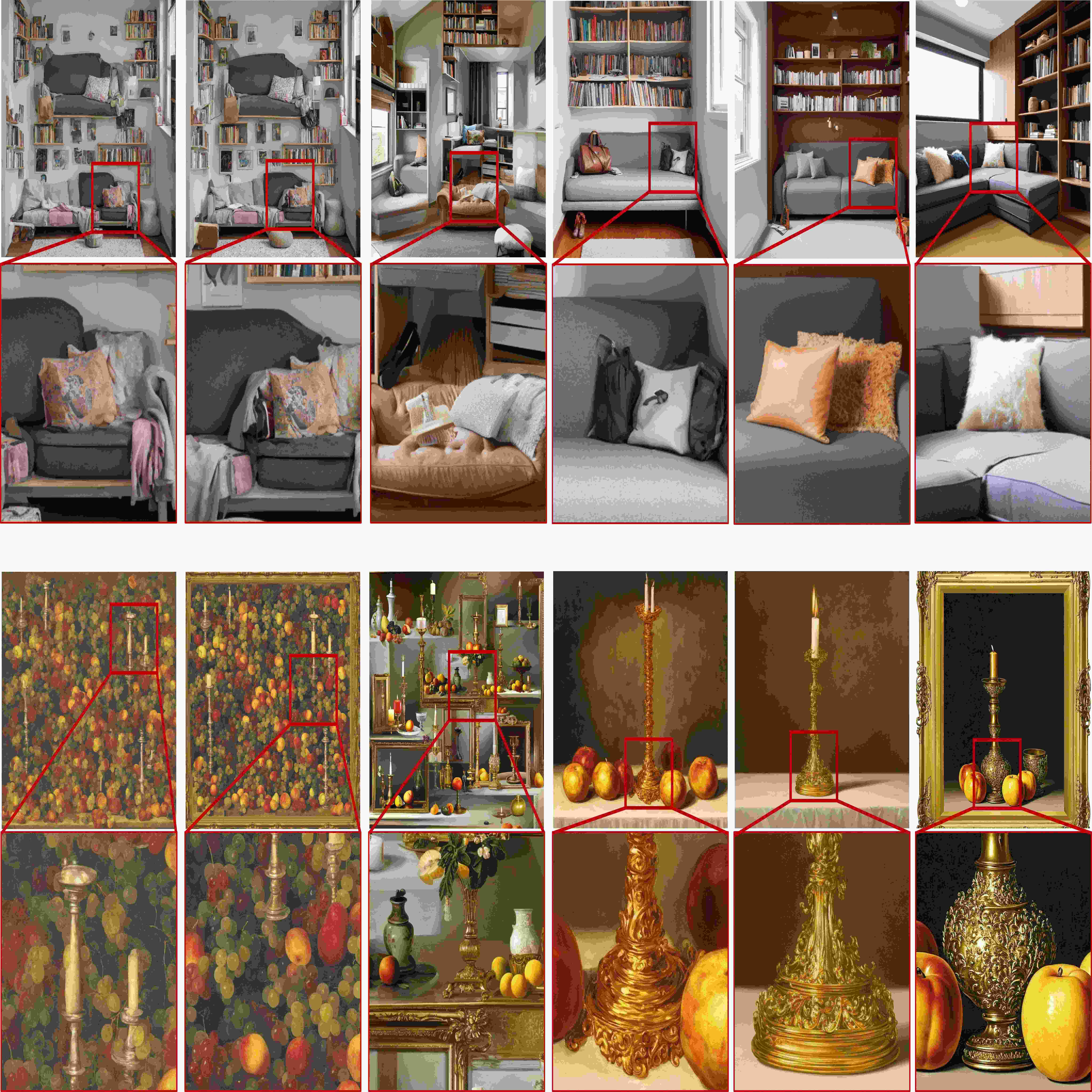}
\put(-3,46){\rotatebox{90}{\footnotesize {$\times 4$ on SD2.1}}}
\put(-3,10){\rotatebox{90}{\footnotesize {$\times 16$ on SD2.1}}}

\put(2,33.5){\scriptsize{DirectInference}}
\put(19,33.5){\scriptsize{MultiDiffusion}}
\put(37.5,33.5){\scriptsize{AttnEntropy}}
\put(54.5,33.5){\scriptsize{ScaleCrafter}}
\put(71,33.5){\scriptsize{HiDiffusion}}
\put(87.5,33.5){\scriptsize{FreCaS(\textbf{ours})}}

\put(2,-1.8){\scriptsize{DirectInference}}
\put(19,-1.8){\scriptsize{MultiDiffusion}}
\put(37.5,-1.8){\scriptsize{AttnEntropy}}
\put(54.5,-1.8){\scriptsize{ScaleCrafter}}
\put(71,-1.8){\scriptsize{HiDiffusion}}
\put(87.5,-1.8){\scriptsize{FreCaS(\textbf{ours})}}
\end{overpic} 

\vspace{14pt}

% \begin{overpic}[width=0.97\linewidth]{figures/appendix_visual_comparsion_sdxl.jpg}
\begin{overpic}[width=\linewidth]{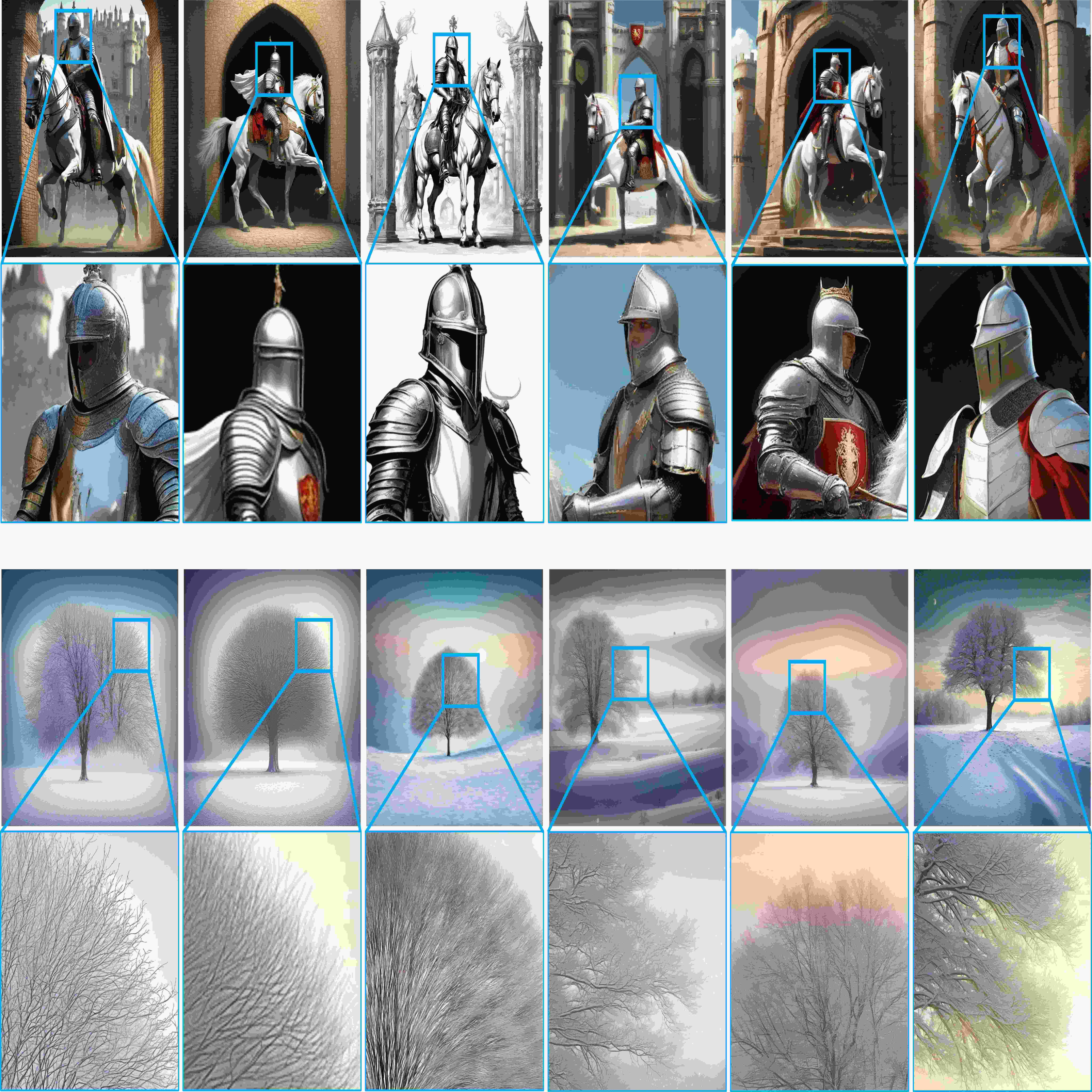}

\put(-3,46){\rotatebox{90}{\footnotesize {$\times 4$ on SDXL}}}
\put(-3,10){\rotatebox{90}{\footnotesize {$\times 16$ on SDXL}}}

\put(3,34){\scriptsize{ScaleCrafter}}
\put(21,34){\scriptsize{FouriScale}}
\put(37.5,34){\scriptsize{HiDiffusion}}
\put(53.5,34){\scriptsize{AccDiffusion}}
\put(70.5,34){\scriptsize{DemoFusion}}
\put(87.5,34){\scriptsize{FreCaS(\textbf{ours})}}

\put(3,-1.8){\scriptsize{ScaleCrafter}}
\put(21,-1.8){\scriptsize{FouriScale}}
\put(37.5,-1.8){\scriptsize{HiDiffusion}}
\put(53.5,-1.8){\scriptsize{AccDiffusion}}
\put(70.5,-1.8){\scriptsize{DemoFusion}}
\put(87.5,-1.8){\scriptsize{FreCaS(\textbf{ours})}}
\end{overpic}

%\vspace{-5pt}

\caption{Visual comparisons on $\times 4$ and $\times 16$ experiments of SD2.1 and SDXL. Please zoom-in for better view.}
\label{fig:appendix_visual_comparisons}

% \vspace{-20pt}

\end{figure*}

\subsection{More Visual Comparisons}
We show more visual comparisons in Figure~\ref{fig:appendix_visual_comparisons}. From top to bottom, the prompts used in the four groups of examples are: 1. ``A small den with a  couch near the window." 2. ``A painting of a candlestick holder with a candle, several pieces of fruit and a vase, with a gold frame around the painting." 3. ``A noble knight, riding a white horse, the castle gates opening." 4. ``Mystical Landscape Digital Art - Lonely Tree Idyllic Winterlandscape by Melanie Viola."

\rebuttal{We have provided more 4K visual comparisons under realistic scenes in Figure~\ref{fig:4k_realstic}. As can be seen, our FreCaS consistently delivers better results in both image layout and semantic details.}
\begin{figure*}
	\centering
	\includegraphics[width=0.98\linewidth]{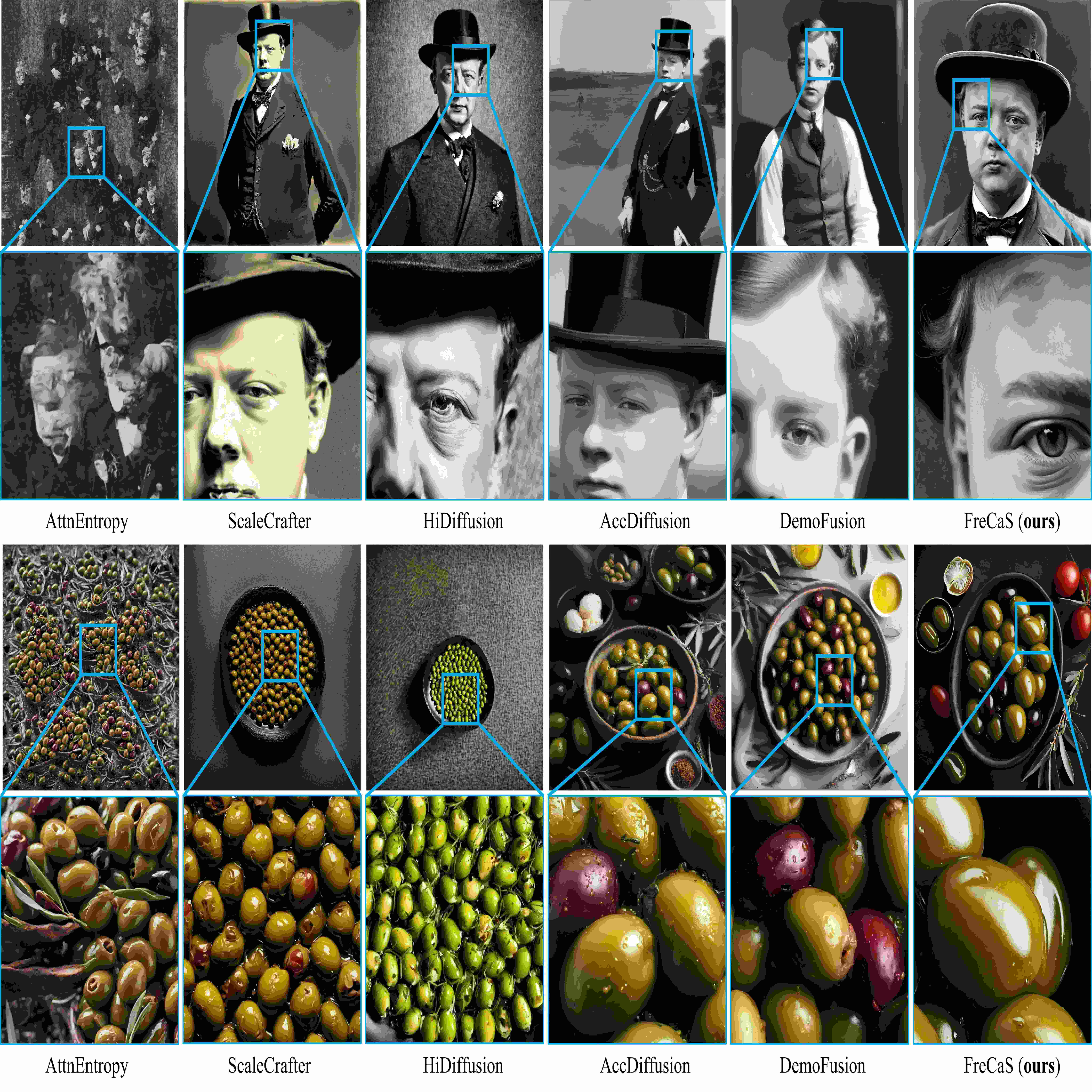}
	\includegraphics[width=0.98\linewidth]{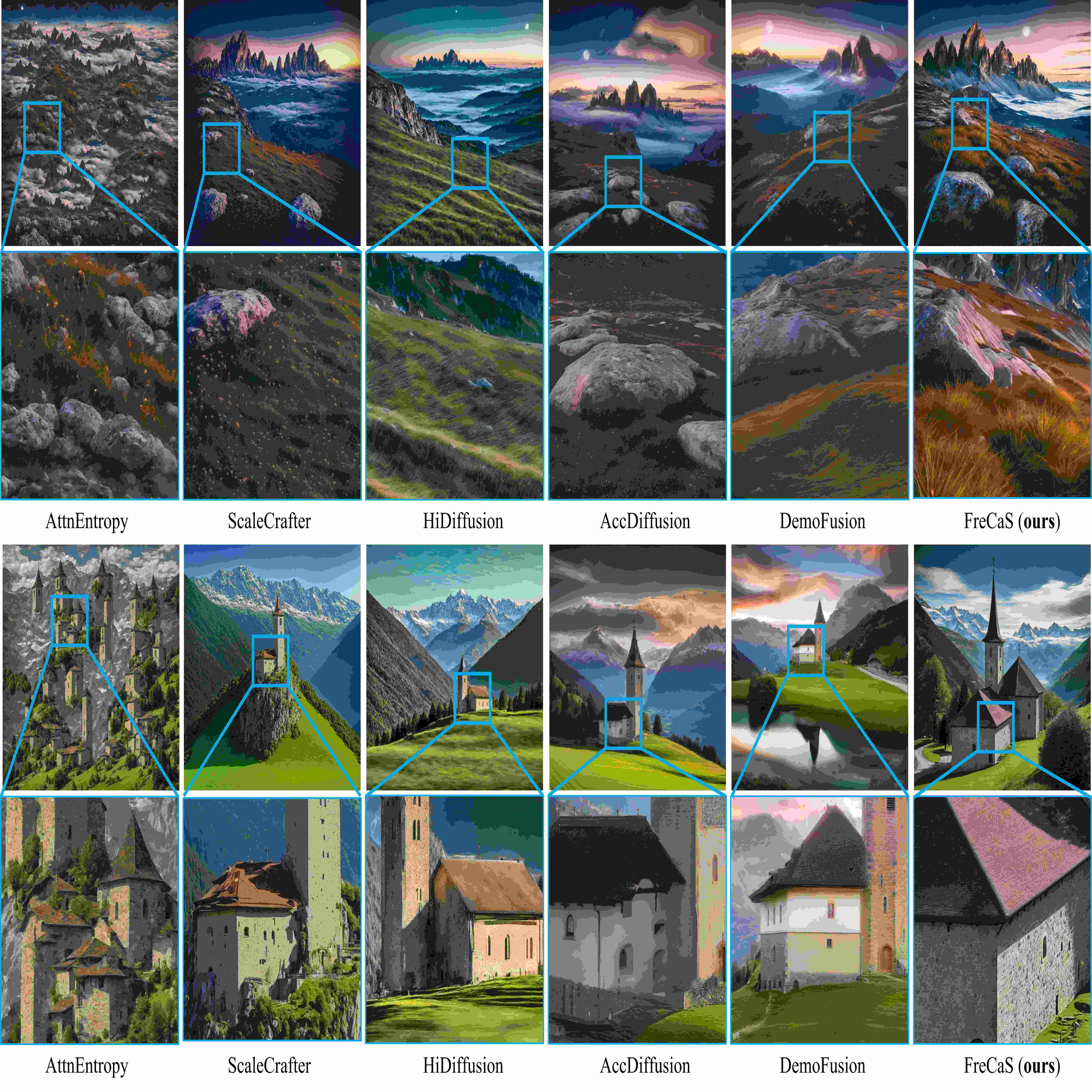}
	\caption{\rebuttal{More 4K comparisons in realistic styles. From top to bottom, the prompts are ``Young winston churchill.", ``Olive food photography.", ``Mountains in fog at beautiful night. Dreamy landscape with mountain peaks, stones, grass, blue sky with blurred low clouds, stars and moon. Rocks at dusk." and ``Image Church Switzerland towers San Romerio Nature Mountains Scenery Made of stone Tower mountain landscape photography."
	}}
	\label{fig:4k_realstic}
\end{figure*}

\section{Experiments on SD3}
\label{appendix_sd3}
\begin{table*}[!t]
    \centering

 \renewcommand{\arraystretch}{1.5}

 \caption{Experiments on $\times$4 generation of SD3.}
 
    \begin{tabular}{c|ccccc|cc}
    \hline
    Methods & $\text{FID}_b$$\downarrow$ & $\text{FID}_p$$\downarrow$ & IS$\uparrow$ & $\text{IS}_p$$\uparrow$ &  \makecell[c]{\footnotesize {CLIP}\\ \footnotesize{SCORE}}$\uparrow$ & Latency (s)$\downarrow$ & SpeedUP$\uparrow$ \\
    \hline
    \hline
    DirectInference & 35.68 & 45.35 & 12.52 & 12.60 & 31.45 & 38.53 & 1$\times$ \\
    Demodiffusion & 15.19 & 44.34 & 17.84 & 14.99 & 31.09 & 63.33 & 0.61$\times$\\
    \colorbox{gray!30}{\textbf{\qquad Ours\qquad}} & 9.76 &  26.62 & 17.83 & 16.72 &31.17 & 15.94 & 2.42$\times$ \\
    \hline
\hline

    \end{tabular}
 \label{tab_appendix_sd3_}

\end{table*}
\begin{figure*}[!t]
%\vspace{10pt}

% \begin{overpic}[width=\linewidth]{figures/sd3.jpg}
\begin{overpic}[width=\linewidth]{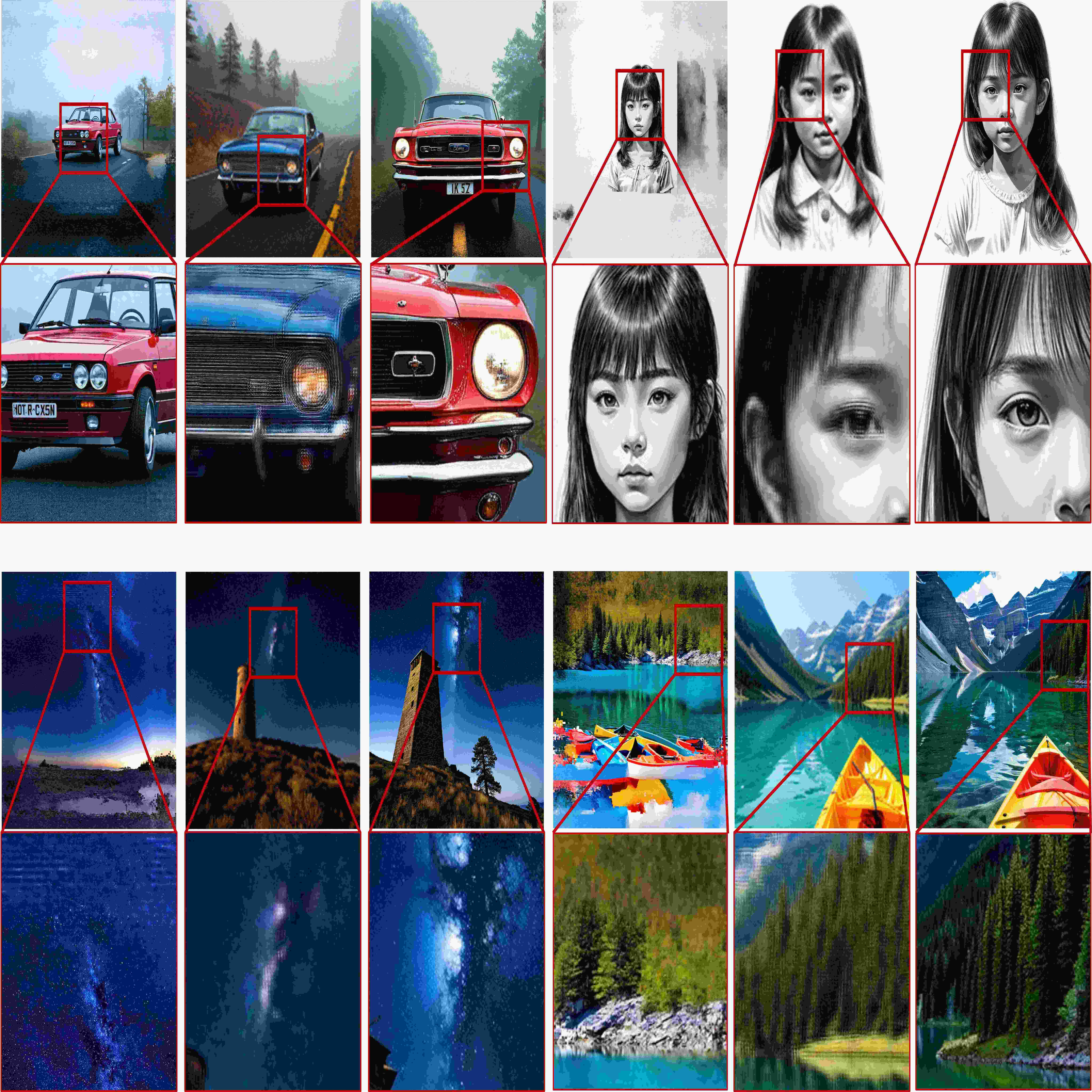}
\put(3,33.8){\scriptsize{DirectInference}}
\put(21,33.8){\scriptsize{DemoFusion}}
\put(37.5,33.8){\scriptsize{FreCaS(\textbf{ours})}}
\put(53.5,33.8){\scriptsize{DirectInference}}
\put(70.5,33.8){\scriptsize{DemoFusion}}
\put(87.5,33.8){\scriptsize{FreCaS(\textbf{ours})}}
	
\put(3,-1.8){\scriptsize{DirectInference}}
\put(21,-1.8){\scriptsize{DemoFusion}}
\put(37.5,-1.8){\scriptsize{FreCaS(\textbf{ours})}}
\put(53.5,-1.8){\scriptsize{DirectInference}}
\put(70.5,-1.8){\scriptsize{DemoFusion}}
\put(87.5,-1.8){\scriptsize{FreCaS(\textbf{ours})}}
\end{overpic}

\caption{Visual comparison on $\times 4$ experiments of SD3. From top to bottom, from left to right, the prompts used in the four groups of examples are: 1. ``Car Photograph - Ford In The Fog by Debra and Dave Vanderlaan." 2. ``Rupert Young is Sir Leon in Merlin season 5 copy." 3. ``Watchtower, Shooting Star \& Milky Way, Gualala, CA." 4. ``Colorful Autumn in Mount Fuji, Japan - Lake Kawaguchiko is one of the best places in Japan to enjoy Mount Fuji scenery of maple leaves changing color giving image of those leaves framing Mount Fuji.". Zoom-in for better view.
	}
\label{fig:sd3}
\end{figure*}

In this section, we present the results of the $\times 4$ generation experiments on SD3. 
SD3 employs a transformer-based denoising network.
It eliminates all convolutional layers, thereby preventing the application of many existing methods, such as ScaleCrafter and FouriScale. 
Besides, SD3 exhibits fine details in the central region but shows corrupted textures in the surrounding regions (see Figure~\ref{fig:sd3}). 
This issue with the image layout also significantly impacts the performance of other methods, such as DemoFusion. 
Therefore, we only compare our FreCaS with DirectInference and DemoFusion.
Table~\ref{tab_appendix_sd3_} and Figure~\ref{fig:sd3} present the quantitative and qualitative results, respectively.

From Table~\ref{tab_appendix_sd3_}, it is evident that FreCaS achieves superior performance in terms of image quality and inference speed.
Specifically, FreCaS achieves the best results on $\text{FID}_b$, $\text{FID}_p$, $\text{IS}$, and $\text{IS}_p$, and only slightly lags behind DirectInference in terms of CLIP score.
Moreover, FreCaS generates a $2048 \times 2048$ image in about 16 seconds, achieving a speed-up of $2.42\times$ and $3.97\times$ compared to DirectInference and DemoFusion, respectively.
Figure~\ref{fig:sd3} illustrates the generated images.
Directly employing the pre-trained SD3 model to generate higher-resolution images, DirectInference leads to unreasonable image layout with the surrounding parts being corrupted, such as the road and trees.
The results of DemoFusion exhibits strange artifacts, such as the car faces and eyes.
In contrast, our FreCaS successfully maintains the natural image structure while obtaining fine details.

\section{Ablation Studies on individual components and Inference Schedule}
\label{appendix_ab}

We further conduct ablation studies to verify the effectiveness of each components and the settings of inference schedule of our FreCaS.

\begin{table}[]
\renewcommand{\arraystretch}{1.3}
\centering
\caption{\rebuttal{Ablation studies on $2048\times 2048$ generation of SDXL.}}
\scalebox{0.9}{
\label{tab:ablation}
    \begin{tabular}{|c|ccc|ccccc|c|}
    \hline
    Model & \makecell[c]{cascaded \\framework} & FA-CFG & CA-reuse & FID$\downarrow$ & $\text{FID}_{p}\downarrow$ & IS$\uparrow$ & $\text{IS}_{p}\uparrow$ & \makecell[c]{\scriptsize{CLIP}\\[-0.3em] \scriptsize{SCORE}}$\uparrow$ & \footnotesize{Latency (s)} \\
    \hline
    \#1 & & & & 39.14 & 29.71 & 11.52 & 14.60 & 32.51 & 34.10 \\
    \cline{5-10}
    \#2 & \cmark & & & 17.62 & 20.49 & 17.01 & 16.54 & 33.24 & 13.71 \\ 
    \cline{5-10}
    \#3 & \cmark & \cmark & & 16.62 & 17.91 & 17.16 & 16.82 & 33.34 & 13.74 \\ 
    \cline{5-10}
    \#4 & \cmark & \cmark & \cmark & 16.48 & 17.91 & 17.18 & 17.31 & 33.28 & 13.84 \\ 
    \hline
    \end{tabular}

    }
\end{table}
\subsection{Effectiveness of each component}

\rebuttal{To better verify the effectiveness of each component of FreCaS, we conducted more ablation studies on our proposed cascaded framework, FA-CFG, and CA-reuse strategies.
The results are shown in Table~\ref{tab:ablation}}.
\rebuttal{One can see that our cascaded framework significantly outperforms the baseline, with a decrease of 22.52 in the FID score and a reduction of 20.39 seconds in latency. This demonstrates the high efficiency of our proposed cascaded framework.
Our FA-CFG strategy improves both FID and IS scores and shows substantial improvement in $\text{FID}_p$, demonstrating its effectiveness in generating realistic image details. The CA-reuse strategy further enhances $\text{IS}_p$,  indicating its effectiveness in improving semantic appearance.
Moreover, these strategies introduce minimal additional latency.}

\subsection{Experiments on Inference Schedule}
\begin{table*}[!t]
	\centering
	
	\renewcommand{\arraystretch}{1.2}

	\caption{Ablation studies on $N$ in FreCaS.}
		\label{tab:appendix_ab1}
	\vspace{5pt}
	
	\begin{tabular}{c|c|cc}
		\hline
		$N$ & resolutions & $\text{FID}_b$$\downarrow$ & $\text{FID}_p$$\downarrow$ \\
		\hline
		0 & 2048 & 43.83 & 29.71 \\
		1 & 1024 $\rightarrow$ 2048 & 12.63 & 17.91  \\
		2 & 1024 $\rightarrow$ 1536 $\rightarrow$ 2048 & 41.36 & 28.68 \\
		\hline
		
	\end{tabular}

\end{table*}

\begin{table*}[!t]
	\centering
	
	\renewcommand{\arraystretch}{1.2}

	\caption{Ablation studies on $L$ in FreCaS.}
		\label{tab:appendix_ab2}
	\vspace{5pt}
	
	\begin{tabular}{c|cc}
		\hline
		$L$ & $\text{FID}_b$$\downarrow$ & $\text{FID}_p$$\downarrow$ \\
		\hline
		0 & 12.57 & 18.20 \\
		100 & 12.69 & 18.10  \\
		200 & 12.63 & 17.91 \\
		300 & 13.30 & 18.57 \\
		400 & 13.34 & 18.62 \\
		\hline
		
	\end{tabular}

\end{table*}
In this section, we conduct experiments on the selection of $N$ (number of additional stages) and  $L$ (the timestep of last latent in each stage).
The two factors are employed to adjust the inference schedule of our FreCaS.
We reports the scores of $\text{FID}_b$ and $\text{FID}_p$ by varying the two factors in Table~\ref{tab:appendix_ab1} and Table~\ref{tab:appendix_ab2}, respectively.

\textbf{Choice of $N$.} 
From Table~\ref{tab:appendix_ab1}, we see that $N=1$ achieves an $\text{FID}_b$ score of 12.63 and an $\text{FID}_p$ score of 17.91, significantly better than $N=0$ and $N=2$ in the $\times 4$ generation task for SDXL. This could be attributed to the fact that a larger value of $N$ introduces more transition steps, which can lead to much information loss. Conversely, a smaller value of $N$ reduces the effectiveness of FreCaS, degenerating it to the DirectInference method.

% In Table~\ref{tab:appendix_ab1}, we can see that $N=1$ achieves a $\text{FID}_b$ of 12.63 and significant outperforms $N=0$ and $N=2$ on $\times 4$ generation of SDXL.
% This is may because a large $N$ leads to enormous transition processes yet much information loss while a small $N$ degenerates FreCaS to conventional DirectInference method. 
% A large $N$ will lead to enormous stages and enlarges the potential information loss during transition, while a small $N$ will contradict to the coarse-to-fine nature of diffusion models and generate similar performance to DirectInference. 
% For the choice of $N$, we find that to make the scale upscaler between adjacent stages as about 2 is proper for existing pre-trained diffusion models. Here, we show the experiment results of SDXL on $\times 16$ generation in Table~\ref{tab:appendix_N}. 
% One can see that if we increase N from 1 to 2 (the count of stage is 3), the overall image quality serious decreases.
% This is mainly due to the potential information loss during the transition process.
% On the other hand, we decrease the N from 1 to 0, our method becomes DirectInference, which has a much lower performance on both details or layout.
% But the same behaviors are also be witnessed in other settings.
% Therefore, we default set the $N$ as 1 for $\times 4$ experiments as well as 2 for $\times 16$ experiments.

\textbf{Choice of $L$.} From Table~\ref{tab:appendix_ab2}, we can see that a smaller $L$ improves $\text{FID}_b$ score but deteriorates $\text{FID}_p$. This is because the details generated at lower resolutions conflict with those at higher resolutions. Thus, we set $L$ to 200 to avoid generating excessive unwanted details in the early stages.

% $L$ determines the degree of content in the low-frequency part. It seems that the best value of $L$ should be about 0 due to the best completation degree.
% However, since the following stages also modify the low-frequency part, the content generated by a too small $L$ sometimes will be contradiction with the new content in the following stage. We present the experiment results at Table~\ref{tab:appendix_L}.
\else
\fi
\end{document}